\pdfoutput=1

\documentclass[11pt]{article}

\usepackage[]{emnlp2021}

\usepackage{times}
\usepackage{latexsym}

\usepackage[T1]{fontenc}

\usepackage[utf8]{inputenc}

\usepackage{microtype}
\usepackage{longtable}
\usepackage{caption}
\usepackage{subcaption}
\usepackage{float}
\usepackage{booktabs}
\usepackage{multirow}
\usepackage{makecell}
\usepackage{graphicx}
\usepackage{colortbl}
\definecolor{mygray}{gray}{.9}
\definecolor{beaublue}{rgb}{0.74, 0.83, 0.9}
\usepackage[shortlabels]{enumitem}

\usepackage{xcolor}
\usepackage{soul}

\newcommand{\ourdataset}[0]{MetaLogic }

\title{
MetaLogic: Logical Reasoning Explanations with Fine-Grained Structure
}

\author{
    Yinya Huang\textsuperscript{1,2}\thanks{~~This work was done when Y. Huang was an intern at Tencent AI Lab.} ~~~
    Hongming Zhang\textsuperscript{2}\footnotemark[2] ~~~
    Ruixin Hong\textsuperscript{3} ~~~
    Xiaodan Liang\textsuperscript{1,4}\thanks{~~X. Liang and H. Zhang are the co-corresponding authors.}~~~ \\
    {\bf Changshui Zhang\textsuperscript{3} ~~~}
    {\bf Dong Yu\textsuperscript{2}} \\
    $^1$Shenzhen Campus of Sun Yat-sen University ~~
    $^2$Tencent AI Lab, Seattle \\
    $^3$Tsinghua University ~~
    $^4$Pengcheng Laboratory \\
    \tt yinya.huang@hotmail.com,
    \tt \{hongmzhang, dyu\}@global.tencent.com, \\
    \tt hrx20@mails.tsinghua.edu.cn,
    \tt zcs@mail.tsinghua.edu.cn, \\
    \tt xdliang328@gmail.com
}

\begin{document}
\maketitle
\begin{abstract}
In this paper, we propose a comprehensive benchmark to investigate models’ logical reasoning capabilities in complex real-life scenarios. 
Current explanation datasets often employ synthetic data with simple reasoning structures.
Therefore, it cannot express more complex reasoning processes, such as the rebuttal to a reasoning step and the degree of certainty of the evidence. To this end, we propose a comprehensive logical reasoning explanation form.
Based on the multi-hop chain of reasoning, the explanation form includes three main components: (1) The condition of rebuttal that the reasoning node can be challenged; (2) Logical formulae that uncover the internal texture of reasoning nodes; (3) Reasoning strength indicated by degrees of certainty.
The fine-grained structure conforms to the real logical reasoning scenario, better fitting the human cognitive process but, simultaneously, is more challenging for the current models. 
We evaluate the current best models' performance on this new explanation form. 
The experimental results show that generating reasoning graphs remains a challenging task for current models, even with the help of giant pre-trained language models. 

\end{abstract}

\section{Introduction}

\begin{figure}[t]
	\centering
	\includegraphics[width=\columnwidth]{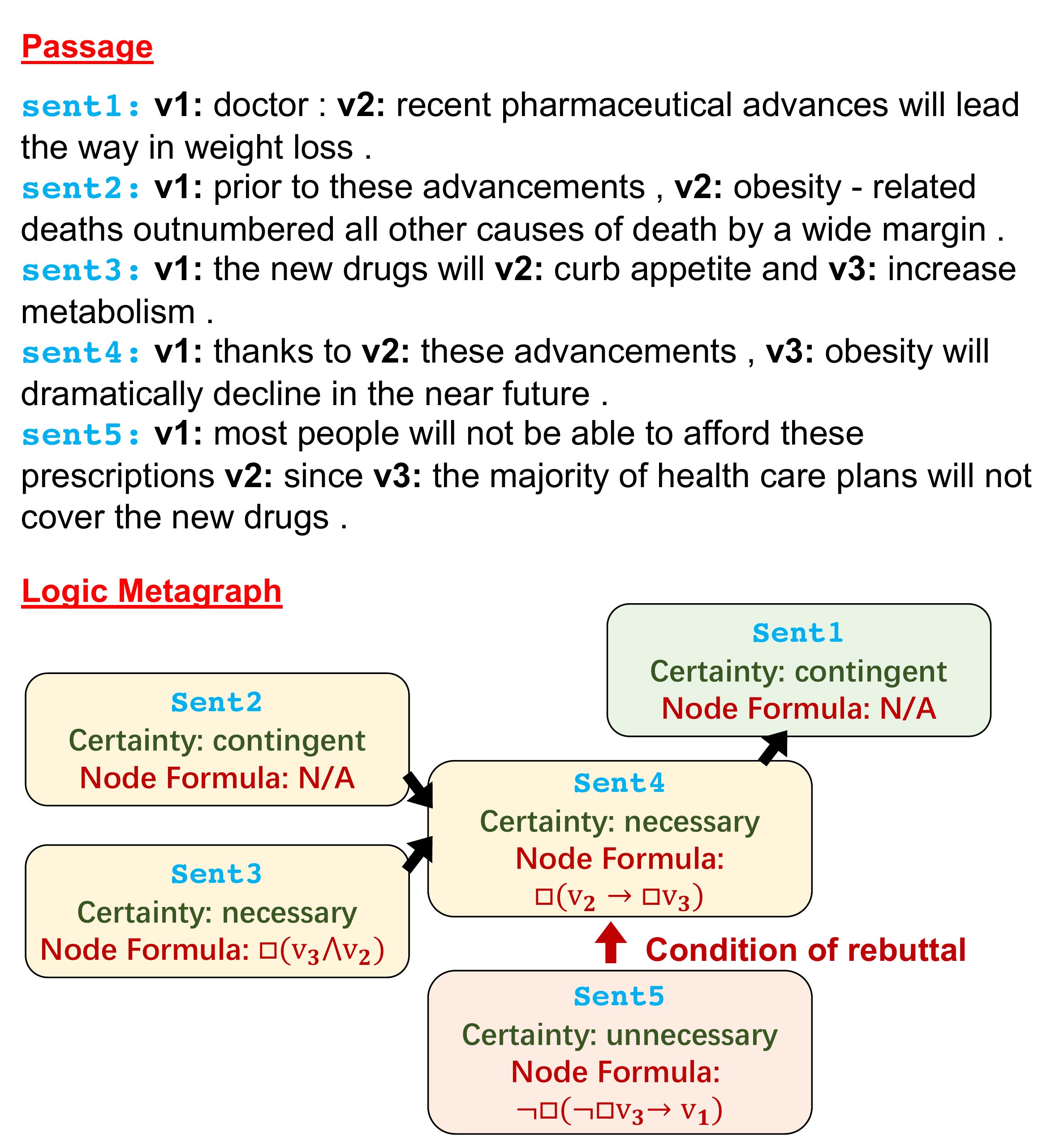}
	\caption{
	\label{fig:intuition}
	A logical passage and the corresponding logic metagraph in the proposed MetaLogic.
	Given a logical passage, 
	the goal is to generate the full metagraph including the chain of reasoning with conditions of rebuttal, the node formulae, and the degrees of certainty. 
	}
\end{figure}

Being able to generate reasonable explanations is a crucial capability for a reliable reasoning system.
Most current works try to ask models to generate reasoning chains as profound explanations.
From simple rationales~\cite{deyoung-etal-2020-eraser} to more complex multi-step explanations~\cite{inoue-etal-2020-r4c,jhamtani-clark-2020-learning,saha-etal-2021-explagraphs} and deductive chains of reasoning~\cite{clark2021transformers,tafjord-etal-2021-proofwriter,dalvi-etal-2021-explaining}, previous works attempt to encompass comprehensive information.
However, the current explanation design still has limitations for logical reasoning texts in real scenarios. 
As current explanations lack a fine-grained structure, three remarkable features are not included in current explanations for the sake of real-world logical reasoning: multiple relation types, hierarchical structure, and certainty. 
As a result, we cannot comprehensively evaluate models' reasoning capabilities in real-life scenarios.

Figure~\ref{fig:intuition} 
shows examples of the crucial reasoning components that are well studied by previous cognitive science literature~\cite{toulmin2003uses,sep-logic-modal} but overlooked by previous work in the machine learning community.
First, the inference \textit{rebuttal}.
Previous work~\cite{tafjord-etal-2021-proofwriter} mostly only focuses on the inferences of \textit{conjunction} and \textit{entailment} among different statements while ignoring the \textit{rebuttal} ones, which could be crucial in real applications.
For example, \texttt{sent5} counters \texttt{sent4} as a condition of exception and we cannot construct the correct reasoning graph without the \textit{rebuttal} relation.
Second, there could exist internal logical relations inside each statement. For example, \texttt{sent5} contains two atomic sentences connected by a logical implication relation.
Third, real-life statements could have different degrees of \textit{certainty}. For example, ``He is hungry'' and ``He is likely to be hungry'' are not identical but relevant because of the certainty.
However, most previous work simply treats them completely separately instead of considering their relevance and trying to model the difference (i.e., certainty).

Motivated by previous cognitive science work (i.e., Toulmin Model\footnote{The Toulmin Model is a canonical theory that helps format and understand arguments. It provides a general pattern to assign logical roles to the sentences in the argument, which clarify the overall logical relations. Especially, the rebuttal components challenge the derivation from existing evidence to the conclusion by providing additional information such as giving a counterexample or proposing an additional condition. }~\cite{toulmin2003uses} and modal logic theory\footnote{The modal logic theory extends classic first-order propositional logic with two modal operators about certainty and several corresponding rules. This facilitates us to keep the logical variables and relations found in the text and, at the same time, introduce degrees of certainty to the graph.}~\cite{sep-logic-modal}), we propose a new explanation form, logic metagraphs, to address the aforementioned limitations of previous work. 
As demonstrated in Figure~\ref{fig:intuition}, the logical metagraphs are directed acyclic graphs with meta nodes connected by two types of edges, \textit{support} and \textit{rebut}, representing the inferences between the statements over a logical passage.
The meta structure uncovers the chain of reasoning from evidence to the conclusion, along with the challenges from the rebuttal sentences. 
Each meta node stores information about a logically sound statement formulated as a propositional formula in a standard modal logic S5 system~\cite{hughes1996new}, a direct extension of first-order propositional logic with two certainty operators. 
The formulae have atomic sentences as logical variables that denote events or beliefs, 
which are modified by three unary operators on their certainty (negation $\neg$, necessity $\Box$, and possibility $\Diamond$) and are joined by three binary operators on their logical relations (implication $\to$, conjunction $\land$, disjunction $\lor$). 
As a result, the logic metagraphs are comprehensive with
multi-hop reasoning paths, inference rebuttal, the internal structure of the statements, and reasoning strength denoted by the degrees of certainty. 
We collect 1,000 logical passages
from the ReClor dataset~\cite{yu2020reclor} and build the \ourdataset dataset.

Based on our new explanation form, we examine the current best models’ ability to understand logical reasoning profoundly.
The models need to generate the logic metagraphs given a logical passage. Performances are evaluated by matching scores for the overall structure as well as the three fine-grained components:
(1) The inference steps between meta nodes; 
(2) The per-statement formulae with multiple logical triples;
(3) The degrees of certainty.
Our evaluation results indicate that generating a comprehensive logical reasoning structure is still challenging for existing giant models.

Our contributions are three-fold:

\begin{enumerate}[leftmargin=*]
	\item We propose a new explanation form, the logic metagraphs, with a comprehensive logical structure and rich logical information, and the corresponding metagraph generation task.
	\item We build a high-quality dataset, MetaLogic, on real-world logical passages.
	\item We conduct experiments on three generative models in different frameworks and locate the challenges for current models.
\end{enumerate}

\begin{figure*}[!t]
    \centering
    \includegraphics[width=16.1cm]{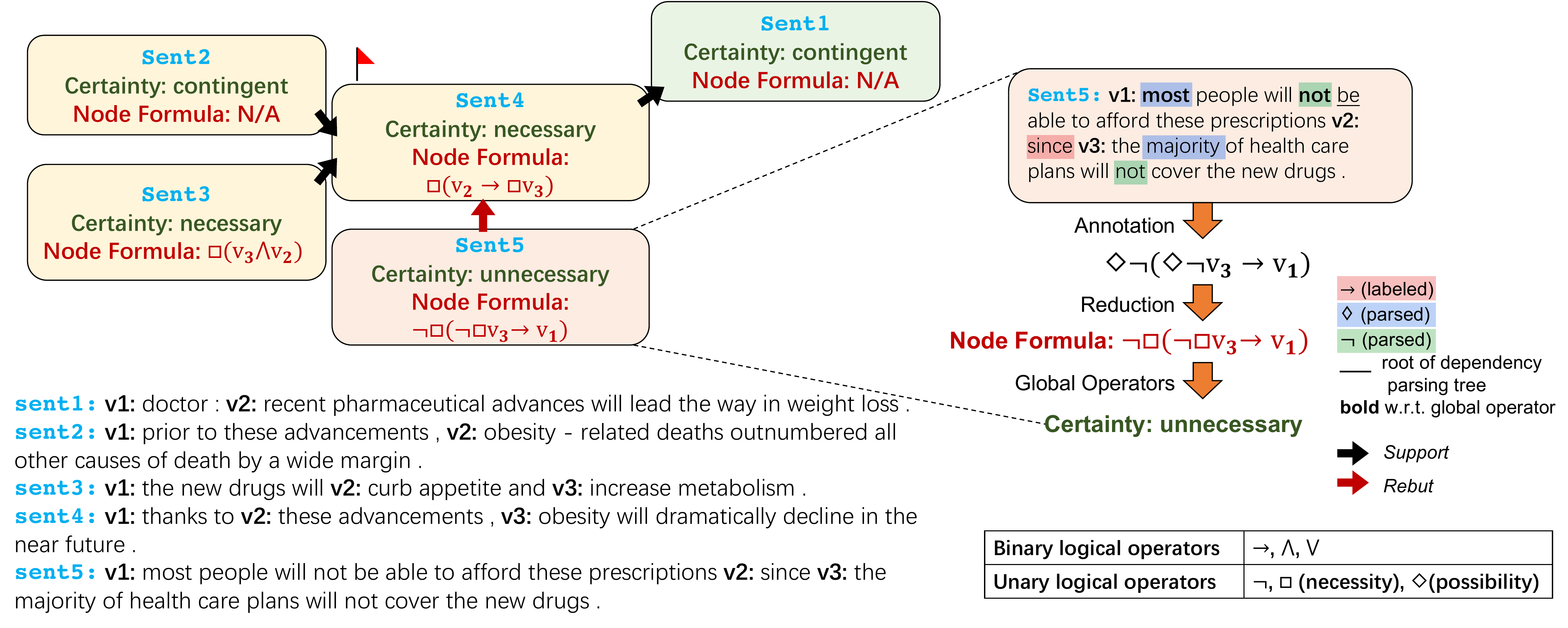}
    \caption{
      \label{fig:trees_overall}
    The overall logical reasoning explanation task is defined as follows. Given a passage, a model reconstructs the fine-grained logical structure with the meta \textit{support} or \textit{rebut} relations, the inner node formulae, and degrees of certainty for each node. Given a logical statement,
    the formula is constructed from the labeled logical triples with the parsed unary operators, which can then be reduced to canonical forms. The certainty label should follow the global operators. 
  }
\end{figure*}

\section{Related Works}
\textbf{Explanations}
Explanation in the context of natural language understanding tasks (e.g., QA) provides interpretability about how models solve the problem. The strategies include asking the models to generate rationales while answering the questions \cite{deyoung-etal-2020-eraser,inoue-etal-2020-r4c}, and deriving multi-hop chains of reasoning \cite{jhamtani-clark-2020-learning,dalvi-etal-2021-explaining}.
The single-sentence rationale provides justification for 
the question answering but does not uncover the reasoning procedure. 
While the form of multi-hop chains of reasoning uncovers the reasoning procedure and remedies the simple justification of rationale, it still lacks critical clues about the mechanism within the reasoning steps. 
Our proposed fine-grained explanation form extends the chain of reasoning by unwrapping the fine-grained texture within each reasoning step. As a result, it allows the reasoning chains to include multiple inference types (e.g., \textit{rebuttal}) and broader reasoning types such as abductive reasoning with the hidden world-knowledge assumption.

\noindent\textbf{Logical Reasoning}
Machine logical reasoning requires models to conduct hidden symbolic reasoning processes through question answering \cite{yu2020reclor,liu2020logiqa,cui-etal-2020-mutual}, or explicitly perform symbolic reasoning via natural language \cite{clark2021transformers,tafjord-etal-2021-proofwriter,dalvi-etal-2021-explaining}. 
The QA-based reasoning data is mostly collected from real-life 
scenarios without corresponding structural information. 
To perform reasoning, 
symbolic modules \cite{huang-etal-2021-dagn,ouyang2021fact} or learning strategies \cite{wang-etal-2022-logic} are designed to approximate the reasoning structure. 
On the other hand, explicitly generating chains of reasoning can better uncover models' reasoning processes. However, recent work mostly focuses on deductive reasoning, 
where models with iterative strategy \cite{tafjord-etal-2021-proofwriter} or reasoning modules \cite{hong2022metgen} show superior performances. 
To encourage more advanced reasoning capabilities, we propose a comprehensive reasoning structure with fine-grained factors.

\begin{table*}[!t]
    \small
	\centering
	\begin{tabular}{
		p{0.03\textwidth}|
		p{0.24\textwidth}|
		p{0.2\textwidth}|
		p{0.18\textwidth}|
		p{0.2\textwidth}
	}
	\toprule
	\rowcolor{beaublue} \multicolumn{5}{c}{Senses} \\
	\midrule
	    & \multicolumn{1}{c|}{Classic} & \multicolumn{1}{c|}{Morality} & \multicolumn{1}{c|}{Tense} & \multicolumn{1}{c}{Belief} \\
	\midrule
	$\Box p$ 
	    & The proposition $p$ is \textit{necessary}. 
	    & $p$ is morally \textit{obligatory}. 
	    & It will \textit{always} be the case that $p$. 
	    & Things a person \textit{knows to be true}. \\
	\midrule
	$\Diamond p$ 
	    & The proposition $p$ is \textit{possible}. 
	    & $p$ is morally \textit{permissible}. 
	    & It will \textit{sometimes} be the case that $p$. 
	    & Things that \textit{may be true} as far as a person knows. \\
	\midrule
	\rowcolor{beaublue} \multicolumn{5}{c}{Definitions} \\
	\midrule
	\multicolumn{2}{c|}{$\Box p := \neg\Diamond\neg p$} & 
	    \multicolumn{3}{c}{It is necessary that $p$. $\quad:=\quad$ It is not possible that not-$p$. } \\
	\multicolumn{2}{c|}{$\Diamond p := \neg\Box\neg p$} & 
	    \multicolumn{3}{c}{It is possible that $p$. $\quad:=\quad$ It is not necessary that not-$p$. } \\
	\midrule
	\rowcolor{beaublue} \multicolumn{5}{c}{Reduction Rules} \\
	\midrule
	\multicolumn{5}{c}{$\Box\neg p = \neg\Diamond p$, \quad $\Diamond\neg p = \neg\Box p$, \quad $\Box\Box p = \Box p$, \quad $\Diamond\Diamond p = \Diamond p$, \quad $\Box\Diamond p = \Diamond p$, \quad $\Diamond\Box p = \Box p$.} \\
	\midrule
	\rowcolor{beaublue} \multicolumn{5}{c}{Degrees of Certainty} \\
	\midrule
    \multicolumn{5}{c}{$\Box := 4$ (necessary), \quad $\Diamond := 3$ (possible), \quad N/A $:= 2$ (contingent), \quad $\neg\Box := 1$ (unnecessary), \quad $\neg\Diamond := 0$ (impossible)} \\
	\bottomrule
	\end{tabular}
	\caption{
		\label{tab:modal}
		Senses, mutual definitions, reduction rules, and degrees of certainty of modal logic operators. 		
	}
\end{table*}

\noindent\textbf{Argumentation / Discourse Structures}
Previous works \cite{lawrence-reed-2019-argument,li2022survey} such as argumentation mining \cite{stab2014identifying,stab2014annotating,stab-gurevych-2017-parsing} or discourse parsing \cite{carlson2001building,webber2019penn} study document structure prediction. Given a passage, a model is required to predict the argument components or the discourse relations between them. Instead of identifying the rhetorical structure of a passage, the proposed logic metagraphs aim at simulating the logical reasoning process, where the model needs to select the relevant knowledge out of a pool to finish the reasoning. 
Besides, unlike directly considering a sentence or a text span as a reasoning node, MetaLogic explores a schema with finer granularity. Each reasoning node is further decomposed into logical variables with relations and modal operators so that the inner structure as well as the certainty are considered.

\section{Task Definition}
\noindent\textbf{Overall Generation Task} The desideratum is that a model reconstructs the fine-grained logic explanation for a given passage, which uncovers the model’s understanding of the logic between the lines.
The logic explanation is formatted as logic metagraphs with \textit{support} or \textit{rebut} inference steps, per-node logical formulae, and degrees of certainty,
as demonstrated in Figure~\ref{fig:trees_overall}. 

The input for the models is a passage with multiple statements $(S^{(0)}, S^{(1)}, ..., S^{(N)})$ and atomic sentences
$p^{(n)}_* \subseteq S^{(n)}$, 
according to which they generate the logic metagraph.
The logic metagraph has three main components: 
(1) The meta structure $\mathcal{G}=(\mathcal{V}, \mathcal{E})$, where $\mathcal{E} = \mathcal{E}_S\bigcup\mathcal{E}_R$, and $\mathcal{E}_S$ and $\mathcal{E}_R$ are the two meta edge types, \textit{support} and \textit{rebut}, respectively, between the meta nodes $u^{(n)}\in\mathcal{V}, n\le N$. 
(2) The set of node formulae $\mathcal{F}$, where
$u^{(n)} := f_n\in\mathcal{F}$.
Each formula is joined by logical triples. $f_n = \bigcap \textsf{r}(\textsf{m}(p^{(n)}_i), \textsf{m}(p^{(n)}_j))$, where $i\neq j$, $\textsf{r}\in \{\to, \land, \lor\}$, and $\textsf{m}$ is a combination in $\{\neg, \Box, \Diamond\}$.
(3) The set of degrees of certainty $\mathcal{C}$, defined by the combination format of $\{\neg, \Box, \Diamond\}$.

\section{The Logic Metagraph}

In this section, we introduce the proposed logic metagraph in details.\footnote{An example is shown on the left side of Figure~\ref{fig:trees_overall}.}

\subsection{Meta Node and Edge}
\label{sec:meta_structure}

Each meta node corresponds to a logically sound statement
(e.g., \textit{premise},
or \textit{conclusion}).
The meta edges are either \textit{support} or \textit{rebut}, relating to a single step of inference. 
The \textit{support} edges join the meta nodes to form a chain of reasoning to the conclusion,
whereas the \textit{rebut} edges indicate challenges from the condition of rebuttal to one of the meta nodes in the chain, which are evidence or claims about exceptional conditions.
Each inference step allows multiple premises.

\subsection{Internal Structure of Meta Node}
\label{sec:meta_node}

The internal structure of a statement is formulated as a propositional logic formula.
The logical variables denote the atomic sentences in the statement that corresponds to separate events or beliefs.
The logical relations between such events or beliefs are denoted by binary propositional operators. There are three logical relations: logical implication, conjunction, and disjunction ($\to$, $\land$, $\lor$). 
Multiple such logical triples are joined by conjunctions ($\land$).
Furthermore, each logical variable and the overall formula are modified by negation ($\neg$) and modal ($\Box$ and $\Diamond$) operators, representing the degrees of certainty of each atomic sentence as well as the whole statement, respectively. 
A more detailed introduction can be found in Section~\ref{sec:modal}.

\subsection{Certainty with Modal Operators}
\label{sec:modal}

Modal logic~\cite{sep-logic-modal} is an extension of first-order propositional logic with two modal operators, 
necessity ($\Box$) and possibility ($\Diamond$).
They are unary operators, and 
Table~\ref{tab:modal} presents examples of their senses in natural language \cite{hughes1996new}. 
For example, $\Box p$ denotes that the proposition $p$ is necessary, while $\Diamond p$ means $p$ is possible, in the classic definition.
In another sense of tense, $\Box p$ represents that the evidence $p$ is true at all times, whereas $\Diamond P$ represents that $p$ is only true sometimes.
In general, the modal operators indicate certainty information of the propositions.

The two modal operators can define each other with the negation operator ($\neg$). Multiple reduction rules are defined. 
As a result, any complex formulae composed of modal operators could be reduced to one of the five degree-of-certainty forms listed in Table~\ref{tab:modal}, which is also known as the classic S5 system~\cite{hughes1996new} and makes the logic metagraph defined in a complete set.

\section{\ourdataset}
\label{sec:annotation}


In this section, we introduce the construction details of the \ourdataset dataset.
Since the logic metagraphs have fine-grained structures with multiple evaluation dimensions, which are all dispensable and supplement each other, we design a rigorous annotation process for the construction. 








\subsection{Preparation}


\textbf{Source Data}
We use ReClor~\cite{yu2020reclor} as the source data,
where the multiple-choice questions are collected from GMAT and LSAT. As a pilot study on logical reasoning explanation, we start with the standard text questions so that the explanation form can benefit from precise and comprehensive logical information.
Each question contains a logical passage, a question, and multiple answer options. 
The original dataset contains 17 reasoning types, which can be mainly categorized into two folds:
complete reasoning composed of the logical passage and the option (e.g., the types Necessary Assumptions, Sufficient Assumptions, Strengthen, Weaken); flawed in-context reasoning structure (e.g., the types Technique, Identify a Flaw, or Dispute). As we aim to study models' understanding of the complete reasoning process over the whole passage, we consider data from the first category, from which we randomly choose 1,000 samples.
Examples of the selected questions can be found in  Appendix~\ref{appx:example_source}.

\noindent\textbf{Data Preprocessing}
We first filter out incoherent options from the questions for logical structure coherence.
For ordinary questions, the incoherent options are the distracting ones.
Conversely, for the inverse questions with ``EXCEPT'', we randomly select one of the distracting options and remove the others. 
We further split the passage into sentences as the initial meta nodes and per meta node sentence into clauses as the initial logical variables. 
This follows the convention of applying linguistic-based segments as reasoning components in related studies \cite{dalvi-etal-2021-explaining,huang-etal-2021-dagn,wang-etal-2022-logic,xu2022logiformer}. Besides, considering the label hierarchy that the logical variables are conditioned on the meta nodes, the initial segments help build the desired metagraph sketch.
Moreover, the initial delimitation is trivial with punctuation marks and provides the least machine guidance to the annotators, who are free to modify the segments on their understanding of reasoning units, which will be demonstrated in Section~\ref{subsec:annotation}.
From the experts' view, 27 of 30 randomly sampled annotated graphs are of high quality, which indicates the high reliability of starting with the initial segments.

As a result, the text presented to the annotators contains the original text with the passage, the question, and the coherent option, along with a list of delimited sentences.

\subsection{Annotation}
\label{subsec:annotation}

As all annotation tasks require a global understanding of the overall passage, we recruit the same annotator to finish all tasks in the same passage.
The annotation procedure has four steps.
(1) Read through the text and have a rough idea about the logical role of each initial meta node (e.g., being a \textit{conclusion} or \textit{rebuttal}).
If an initial meta node does not provide complete evidence, then the annotator needs to merge it with another node to form complete evidence. (2) Annotate the inference types between the meta nodes. After this stage, we obtain the chain of reasoning and the rebuttal steps. 
(3) For each meta node, annotate the logical variables by refining the span boundaries of the given initial logical variables. (4) Annotate the logical binary operator between the logical variables. 
The annotation platform is demonstrated in Appendix~\ref{appx:annotation_details}.

\begin{table}[!t]
    \small
    \centering
	\begin{tabular}{
        ccccc
	}
	    \toprule
         & \multicolumn{2}{c}{Meta Structure} & \multicolumn{2}{c}{Meta Node} \\
        & M-Node & M-Edge & L-Variable & L-Relation \\
		\midrule
        $\mathsf{\kappa}$ & 57.80$^{\dag}$ & 42.82$^{\dag}$ & 65.46$^{\ddag}$ &   56.81$^{\dag}$  \\ 
		\bottomrule
	\end{tabular}
	\caption{
		IAA with Cohen's Kappa coefficients. 
		M-Node: meta node, M-Edge: meta edge, L-Variable: logical variable, L-Relation: logical relation. 
		$^{\ddag}$ indicates very high agreement with $\kappa$ over $60\%$.
		$^{\dag}$ indicates high agreement with $\kappa$ between $40\%$ and $60\%$.
	}
	\label{tab:iaa}
\end{table}

We recruit annotators from crowd-sourcing platforms.
We first train annotators with a carefully designed annotation guideline\footnote{Details are shown in Appendix~\ref{appx:map_binary_operators}.} and require them to pass an exam before the annotation to guarantee the annotation quality.
For each passage, we invite two annotators\footnote{For the inconsistent annotation, we invite a third annotator to make the judgement.}.
On average, we pay \$2.2 for each logical passage. 



\begin{table*}[!t]
  \small
  \centering
  \begin{tabular}{
    c|cccc|cc|ccc|c
  }
    \toprule
   & \multicolumn{4}{c|}{Component 1} & \multicolumn{2}{c|}{Component 2} & \multicolumn{3}{c|}{Component 3} & \multirow{2}*{Overall} \\
   & \multicolumn{2}{c}{Node} & \multicolumn{2}{c|}{Step} & \multicolumn{2}{c|}{Formula} & \multicolumn{3}{c|}{Certainty} & \\ 
   & F1 & All & F1 & All & F1 & All & Acc & All & F1${^*}$ & All \\ \midrule
  Once (large) & 91.0 & 55.7 & 45.3 & 15.0 & 58.8 & 57.8 & 79.2 & 40.8 & 24.3 & 4.4 \\ 
  Multitask (large) & 92.8 & 62.5 & 52.3 & 20.7 & 77.2 & 75.8 & 82.8 & 52.0 & 41.5 & 8.9 \\ 
  MetGen (large) & 94.5 & 68.7 & 56.3 & 22.3 & 78.3 & 76.7 & 84.0 & 56.3 & 50.3 & 11.2 \\
  \midrule
  Once (11b) & 92.5 & 59.0 & 52.4 & 23.5 & 71.0 & 69.6 & 84.1 & 55.0 & 37.2 & 11.7 \\
  Multitask (11b) & 93.9 & 66.5 & 58.3 & 28.0 & 77.8 & 75.7 & 82.8 & 54.5 & 45.6 & 13.2 \\
  MetGen (11b) & 94.4 & 68.5 & 61.5 & 28.0 & 80.8 & 78.8 & 85.7 & 60.0 & 55.7 & 15.4 \\
  \bottomrule
  \end{tabular}
  \caption{
    \label{tab:exp}
    Evaluation results of generative models. 
    All: AllCorrect. $^*$: macro-F1. 
  }
\end{table*}

For unary logical operators ($\neg$, $\Box$, $\Diamond$), 
as discussed by~\cite{toulmin2003uses}, there exist conventional clue words for the negation and modality. 
Following that, we leverage such in-context clue words for the annotation.
%
Given a set of conventional indicators (demonstrated in Table~\ref{tab:indicators} in Appendix~\ref{sec:appendix_indicators}), we parse each meta node sentence into a dependency parsing tree, then detect those words within 3-hops to the root node, and assign the corresponding operators to the formula. 
The consecutive unary operators are ordered by the distance from the indicators to the parsing root node. This results in the global unary operators.
%
For local unary operators of the logical variable spans, we parse the spans and evaluate the indicator-root distance.
The repeatedly detected indicators are reduced, as a result, the operators are kept by the global formula and removed before the local variable. 
%
To evaluate the labels, the annotators check 391 unary operators from 200 randomly sampled passages. 
As a result, 92.6\% of them are consistent with human cognition, which indicates that the operators are valid and consistent.






\subsection{Inter-Annotator Agreement}

We evaluate the inter-annotator agreement in multiple dimensions with Cohen's Kappa Coefficient~\cite{cohen1960coefficient}.

\noindent\textbf{Meta Node}
The IAA of meta nodes reflects one's understanding of the logical role of each statement. 
We evaluate the annotators' agreement of each meta node of being one of the five characters: conclusion, rebuttal, beginning of the chain, an intermediate conclusion in the chain, and irrelevant node.

\noindent\textbf{Meta Edge} 
We consider the exhaustive meta node pairs except the reflexive ones.
Consequently, the agreement is calculated on the adjacency matrix of the meta edges regarding the three labels: \textit{support}, \textit{rebut}, and without-an-edge. The diagonal elements in the matrix are excluded.

\noindent\textbf{Logical Variable} 
As the logical variables are text spans, the annotators vote for each token for being in a logical variable or not. The agreement is average over the per-token agreement. 

\noindent\textbf{Logical Relation}
Similar to meta edge, we consider the exhaustive logical variable pairs except for the reflexive ones. Considering the logical variables as vertices, the agreement is calculated on the adjacency matrix regarding the four labels: logical implication, logical conjunction, logical disjunction, and without-a-relation. The diagonal elements are regarded.

We present the results in Table~\ref{tab:iaa}.
The agreement is consistently high, which indicates the high quality of MetaLogic. Moreover, the high IAA also indicates that humans could easily solve the logical reasoning explanation and provide consistent logical reasoning graphs.

\subsection{Dataset Statistics} 
The final annotated \ourdataset contains 1,000 logic metagraphs with over 3,609 meta nodes and 1,500 formulae. In the MetaLogic, 416 out of 1,000 logic metagraphs have rebuttal steps. Around 40\% of metagraphs have multi-hop reasoning chains. On average, each logic metagraph has more than three meta nodes, and about 40\% are mapped to formulae. Moreover, each metagraph has an average of 2.19 global operators.
More statistics can be found in Table~\ref{tab:stat}. 
We randomly split the data with 60\% training, 20\% development, and 20\% testing.

\begin{table}[!t]
  \small
  \centering
  \begin{tabular}{
    p{.08\columnwidth}
    p{.08\columnwidth}
    p{.08\columnwidth}
    p{.08\columnwidth}
    p{.08\columnwidth}
    p{.08\columnwidth}
    p{.1\columnwidth}
  }
    \toprule 
    Passage & Graph & Node & Form & Reb & Multi-step & Multi-premise \\
    \midrule 
    1,000 & 1,000 & 3,609 & 1,500 & 416 & 435 & 400 \\    
    \midrule \midrule
    \multirow{2}*{\makecell[c]{Avg.\\Node}} & \multirow{2}*{\makecell[c]{Avg.\\Form}} & \multirow{2}*{\makecell[c]{Avg.\\Var}} & \multirow{2}*{\makecell[c]{Avg.\\Binary}} & \multicolumn{2}{c}{\makecell[c]{Avg. Unary}} & \\
     & & & & Global & Local \\
    \midrule
    3.61 & 1.5 & 2.63 & 1.89 & 2.19 & 1.31 & \\
    \bottomrule 
  \end{tabular}
  \caption{
    \label{tab:stat}
    Label statistics.
    The first row indicates the overall numbers of passages (passage), logic metagraphs (graph), meta nodes (node), formulae (form), number of metagraphs that have \textit{rebuttal} steps (reb), have multi-step chains (multi-step), and have multiple premises (multi-premise).
    Note that reb, multi-step, and multi-premise have small intersections and $\textsf{n}_{\textrm{reb}\cap \textrm{step}\cap \textrm{premise}} = 36$.
    The second row shows per-passage average number of meta nodes, formulae, logical variables (var), binary operators (binary), global and local unary operators (global, local).
  }
\end{table}

\section{Experiment}

We evaluate the performance of the following explanation generative models on MetaLogic: (1) All-at-Once (\textbf{Once}) T5~\cite{raffel2020exploring}, which performs sequence-to-sequence generation via generating the whole metagraph in a linearized sequence given the overall passages with the sentence and variable denotations; (2) \textbf{Multitask} T5~\cite{raffel2020exploring}, which complete the whole generation task with a combination of three sub-tasks: meta structure generation, formula generation, and certainty prediction; (3) \textbf{MetGen}~\cite{hong2022metgen}, which is a module-based framework for structured explanation generation.
It further introduces a reasoning controller and two modules for meta structure generation.
To the best of our knowledge, MetGen is the current state-of-the-art explanation generative model.
Further model details are in Appendix~\ref{appx:model_details}.
Following~\citet{dalvi-etal-2021-explaining}, we report the F1 and AllCorrect scores for each dimension and the overall AllCorrect score.
For certainty, we report the accuracy of a five-label classification and an extra macro-F1 due to the unbalance of the degree labels. 
The overall AllCorrect is the strictest metric since any difference in the predicted metagraph will make the prediction a wrong one.
Details can be found in Appendix~\ref{sec:evaluation_metrics}.

\begin{figure}[!t]
    \centering
    \includegraphics[width=7.5cm]{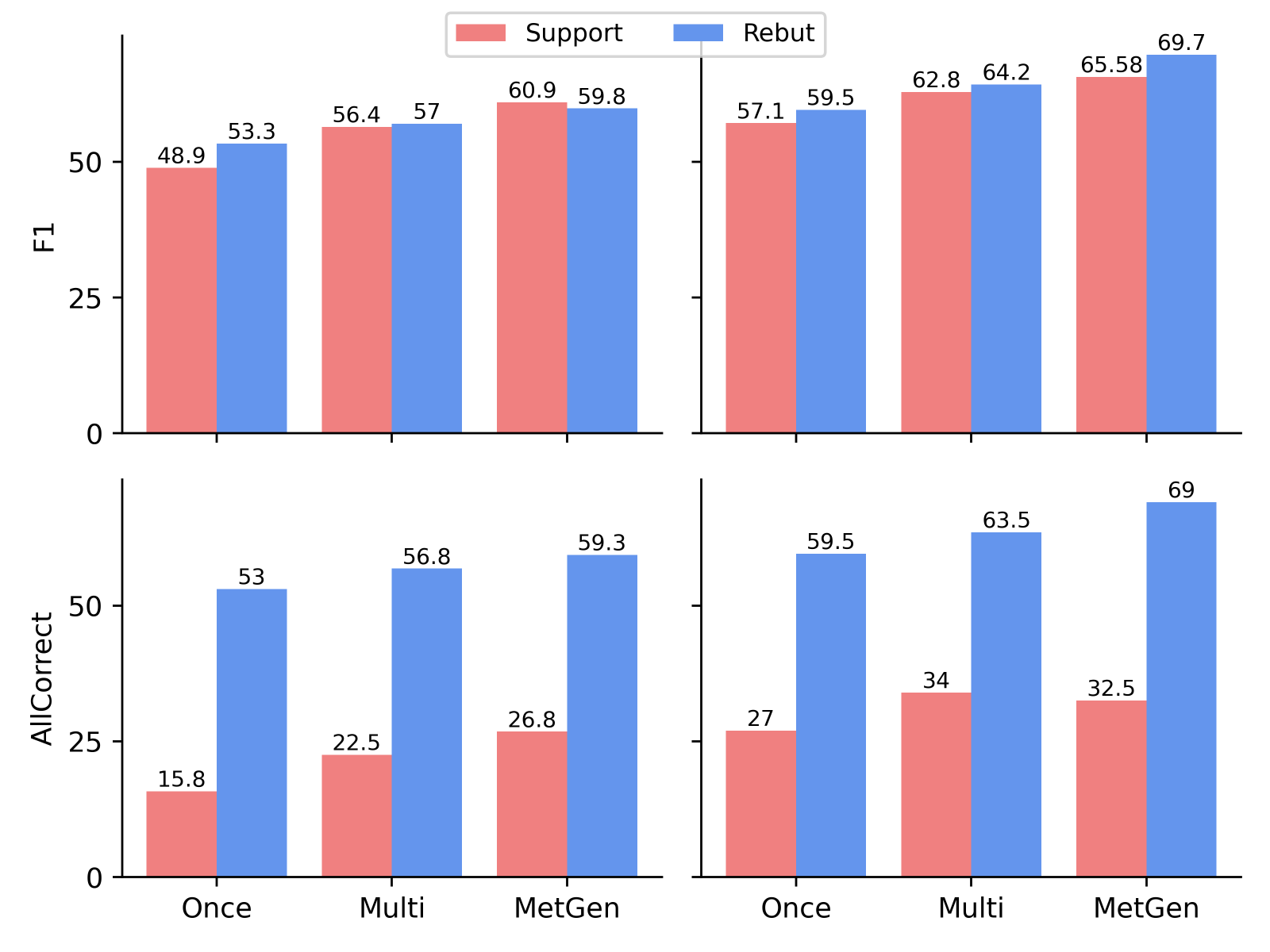}
    \caption{
        Performances on different inference types: \textit{support} and \textit{rebut}.
        Left column: models with the T5-large backbone. Right column: models with the T5-11b backbone.
    }
    \label{fig:step}
\end{figure}

\subsection{Implementation Details}
We fine-tune Once (large), Multitask (large), and MetGen (large) with a batch size of 32 for 300 epochs on 1 Tesla V100 GPU, and fine-tune Once (11b), Multitask (11b), MetGen (11b) with a batch size of 4 for 300 epochs on 8 Tesla V100 GPUs. The learning rate is 1e-5 for all models. The model parameters are optimized by Adafactor~\cite{shazeer2018adafactor}. The models are evaluated per 10 epochs on the development set, and the best checkpoints are saved for test set evaluation.

\begin{figure}[!t]
    \centering
    \includegraphics[width=8cm]{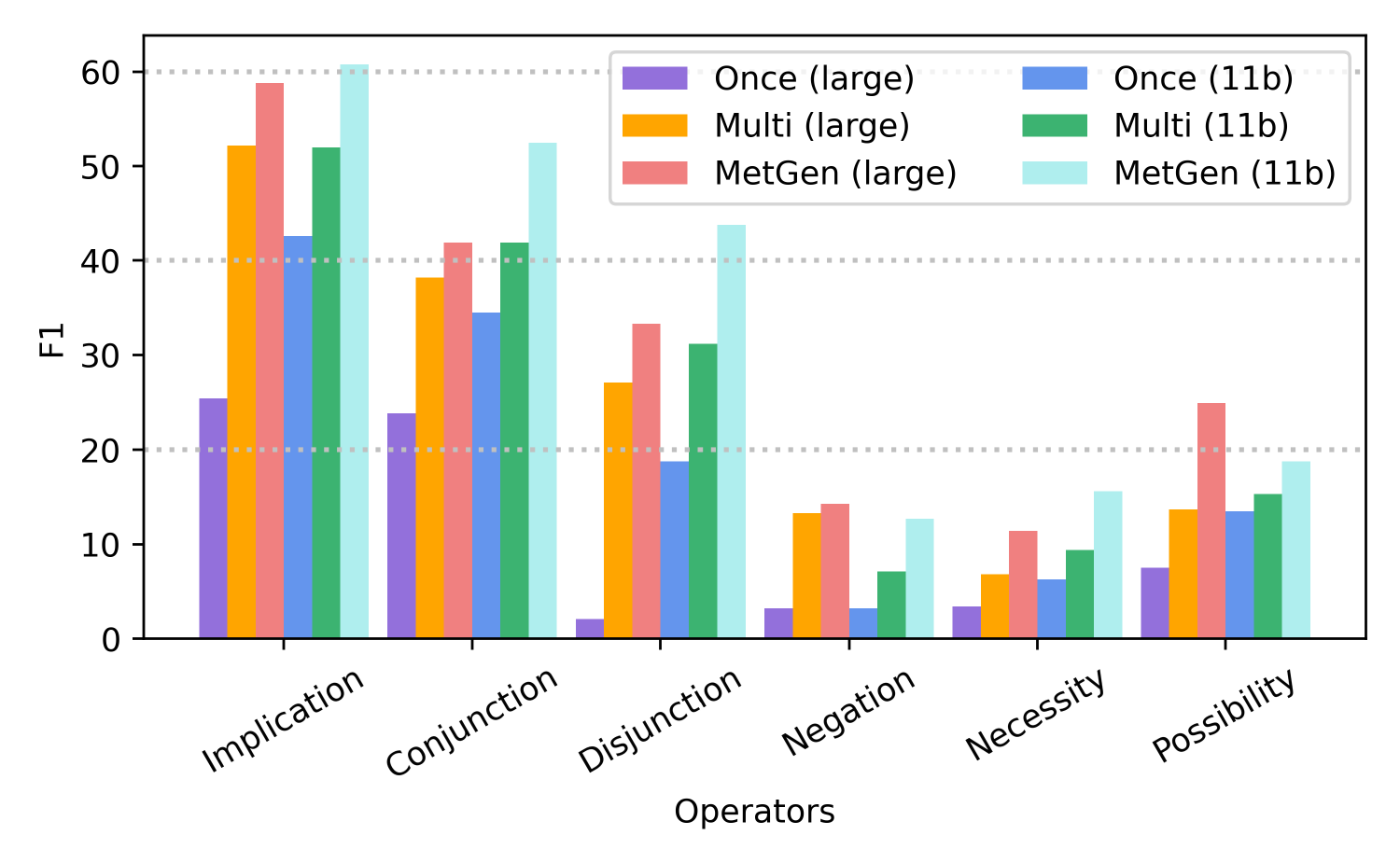}
    \caption{
        Performances on each logical operator.
    }
    \label{fig:per_operator}
\end{figure}

\subsection{Main Results}
From the results shown in Table~\ref{tab:exp}, we can find out that generating the comprehensive logic graph is still a very challenging task, even for current giant models as all models achieve low AllCorrect performance.
Specifically, we can make the following observations:
\begin{enumerate}[leftmargin=*]
    \item From the experiments in certainty prediction, we can see that all models are struggling, which shows that knowing certainty is still not a trivial task for current models given that there are explicit indicators in context. 
    \item We notice that using larger pre-trained models (e.g., T5-11B) can help improve the performance of all models, this indicates that big models can help better model the statement semantics such that they can better identify and link statements.
    \item We also notice that the module-based method MetGen can outperform the Once and Multitask method, which indicates that iteratively generating the explanation graph with basic modules is a more reliable logical reasoning framework.
    \item From the experiments on Component 1, we can see that the models could obtain high node scores and mediocre step scores, but the step AllCorrect results are inferior. This indicates that with the help of giant pre-trained LMs, current models could effectively learn to identify the nodes, but they may not know the true logical reasoning because they cannot precisely predict the inference types among these nodes.
    \item The models achieve around or over 60\% of F1 and AllCorrect scores in predicting the formula, showing their awareness of the inner logical structure. This makes sense because the majority of the inner structure is triggered by connective words such as ``so.''
    
\end{enumerate}

In the rest of this section, we present a more detailed inspection from different perspectives.

\begin{figure*}[!t]
    \centering
    \includegraphics[width=\textwidth]{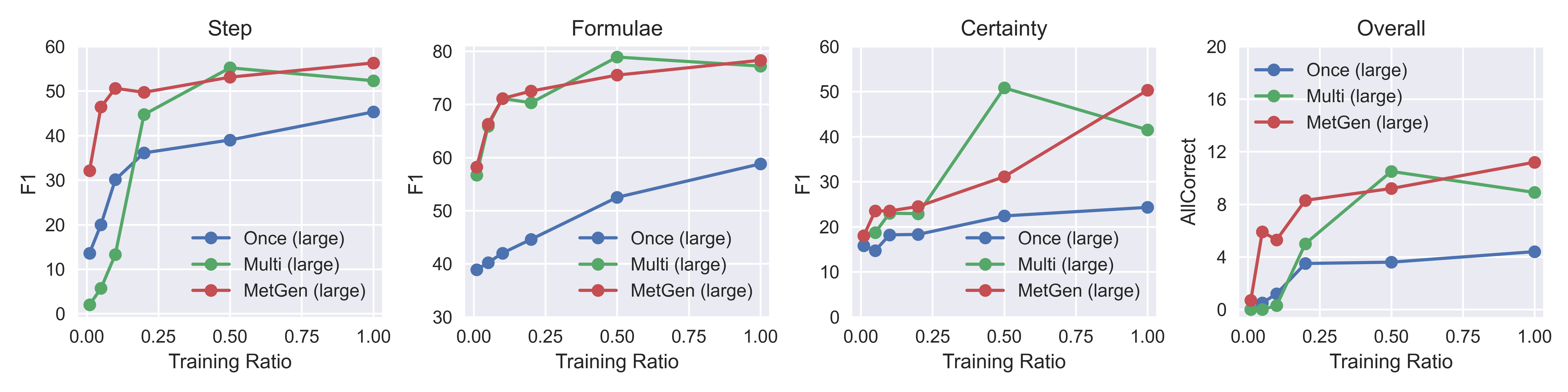}
    \caption{
        Results on different ratios (0.01, 0.05, 0.1, 0.2, 0.5, 1.0) of \ourdataset training data.
    }
    \label{fig:few_shot}
\end{figure*}

\subsection{Performances on Metagraph Parts}
We further inspect models' performance on detailed components.
The evaluation results on separate inference types (\textit{support} and \textit{rebut}) are demonstrated in Figure~\ref{fig:step}. 
Overall, identifying \textit{rebut} is easier than \textit{support}, according to the exact match scores F1 and AllCorrect.
This makes sense because most \textit{rebut} nodes could contain informative keywords such as ``however'' but the majority of nodes in \textit{support} edges do not.
Besides, the average F1 scores per operator are shown in Figure~\ref{fig:per_operator}.
From the results, we can see that the trend of models' performance are generally consistent on different operators, which indicates that different operators may have different intrinsic difficulty.

\subsection{Data Scale for Logical Inference}

To investigate how well current models can learn to generate the reasoning graphs, We use different ratios of training data to train the models and present the results in Figure~\ref{fig:few_shot}.
Overall, the model performances show a rapid increase within 20\% of training data, then a flat and steady increase and do not reach a platform, indicating that the models can still benefit from more structural reasoning data. Among the models, MetGen has the most significant growth trend and performs data efficiently with small data, showing the advantages of the module-based learning framework in symbolic reasoning. 
Interestingly, we find out that the performance of multitask T5 decreases after seeing half of the training data. A possible explanation is that the decomposed logical structure as independent sub-tasks prevents the models from a holistic understanding of the logical passages.
Besides that, the flat increasing rate after seeing 20\% of the training data also suggests that blindly increasing the training data scale may not be the most efficient way of teaching models to conduct such a complex reasoning task.

	


\subsection{Error Analysis}

\begin{table}[]
    \small
    \centering
    \begin{tabular}{ccc}
         \toprule
         Graph & Formula & Certainty \\
         (G1-G5) & (F1-F4) & (C1-C4) \\
         \midrule
         13/13/11/10/12 & 32/6/8/4 & 5/30/10/5 \\
         \bottomrule
    \end{tabular}
    \caption{
        \label{fig:error_type}
    Error type statistics. We randomly select 50 test set predictions and group the error types by components. 
    G1: Incorrect inference type. 
    G2: Incorrect rebuttal.
    G3: Incorrect conclusion.
    G4: Incorrect inference step.
    G5: Other structural mismatch.
    F1: Incorrect logical variable.
    F2: Incorrect unary operator. 
    F3: Incorrect binary operator.
    F4: Incorrect implication direction.
    C1: Incorrect polarity.
    C2: Other polarities to contingent.
    C3: Contingent is predicted as other polarities.
    C4: Unresolved degree of certainty.
    }
    \label{tab:my_label}
\end{table}

To better understand current models' errors, we randomly sample 50 instances from the development set and collect the predictions from the All-at-Once (T5-11b) model. 
We manually evaluate the predictions and categorize 4 to 5 error types for each component, as shown here in the Table~\ref{fig:error_type}. 

Specifically, the meta graph structure mainly has five error types: (G1) Incorrect inference type: the model predicts the correct structure, but over one of the inference steps has the inverse type (i.g., predicted \textit{support} but should be \textit{rebut} or vice versa); (G2) Incorrect rebuttal: missing or incorrectly predict a rebuttal step; (G3) Incorrect conclusion: mismatched conclusion node at the end of the reasoning chain; (G4) Incorrect inference step: missing or predicting redundant inference step; (G5) Other structural mismatches: Including different chain branches and so forth.
The four error types in formulae are (F1) Incorrect logical variable: missing or predicting redundant logical variable, or predicting a wrong variable; (F2) Incorrect unary operator: The variables are correct but are bound by incorrect unary operators; (F3) Incorrect binary operator: The variables and unary operators are correct, but predict incorrect binary operator. (F4) Incorrect implication direction: The variables, unary and binary operator types are correct, but the implication operator has an inverse direction. 
The four error types in certainty: (C1) Incorrect polarity: Predicting the certainty in an opposite polarity; (C2) Other polarities to contingent; (C3) Contingent is predicted as other polarities; (C4) Unresolved degree of certainty. 

From the results we can see that, the model tends to predict the operator quite well (F2/F3), but not the variable (F1), which suggests that even though current deep models can identify the correct relations with some trigger words (e.g., ``so that''), they may not fully understand it because they cannot find the correct variable span in the context.
Besides that, we also notice that the model tends to predict the wrong polarities, which is typically irrelevant towards the conclusion, as the important certainty feature. This suggests that the model may learn to answer questions with the wrong reason (i.e., short path~\cite{DBLP:conf/iclr/LoveringJLP21}), which further demonstrates the importance of our task for constructing a reliable and trustworthy reasoning system.

\section{Conclusion}
This paper extends the boundary of current research on logical graph generation for reliable reasoning systems.
Specifically, we carefully design a complete logic explanation form following previous research on cognitive science.
Accordingly, we built \ourdataset with a comprehensive annotation task design and quality examination.
We also evaluate several recent models and show that the performance of current models is still unsatisfactory, even with giant pre-trained language models. 
We hope that this paper could motivate more future works on reliable reasoning systems that could generate the correct logical graphs to support their reasoning.
The MetaLogic data and implementation code are available at \url{https://github.com/tencent-ailab/MetaLogic}.


\section{Limitation}

The major limitation of \ourdataset is that we cannot annotate a large enough dataset for data-driven methods. However, considering that humans could learn to conduct logical reasoning after seeing a few examples, we argue that it is meaningful to investigate whether machines can learn the same level of reasoning capability with limited data.

\section{Ethical Considerations}
During the annotation process, we follow the minimum payment requirement of the united states. No personal or confidential information is collected. Hence, to the best of our knowledge, there is no ethical concern.

\section*{Acknowledgements}
We appreciate the anonymous reviewers for their insightful comments. We thank Dr. Su Wu for reviewing the annotation manual, thank Dr. Xingchi Su for double-checking the operator reduction, and thank Jianheng Tang, Zhicheng Yang, and Xinran Zhao for their constructive advice for the manuscript.
This work was supported in part by National Key R\&D Program of China under Grant No. 2020AAA0109700, National Natural Science Foundation of China (NSFC) under Grant No.61976233, Guangdong Province Basic and Applied Basic Research (Regional Joint Fund-Key) Grant No.2019B1515120039, Guangdong Outstanding Youth Fund (Grant No. 2021B1515020061), Shenzhen Fundamental Research Program (Project No. JCYJ20190807154211365) and CAAI-Huawei MindSpore Open Fund. We thank MindSpore for the partial support of this work, which is a new deep learning computing framework\footnote{https://www.mindspore.cn/}.

\bibliography{anthology,custom}
\bibliographystyle{acl_natbib}

\newpage
\appendix

\section{Example Source Data from ReClor}
\label{appx:example_source}

Tables~\ref{tab:reclor_necessary_assumption}, \ref{tab:reclor_sufficient_assumption}, \ref{tab:reclor_strengthen}, and \ref{tab:reclor_weaken} demonstrate the example source data of different reasoning types from the ReClor dataset.


\begin{table}[b!]
    \small
    \centering
    \begin{tabular}{p{.95\columnwidth}}
        \toprule
        \textbf{Reasoning Type: Necessary Assumptions} \\
        \midrule
        \textbf{Context: } A recent study showed that people who address problems quickly and directly are significantly less likely to have gum disease than are people who react to problems by refusing to think about them. Since stress can have a negative effect on the immune system, the study' s results clearly indicate that some forms of gum disease are caused or aggravated by suppression of the immune system. \\
        \textbf{Question: } The argument requires the assumption that \\
        \textbf{Options: } \\
        A: people who tend to address problems quickly and directly will invariably seek dental care at the first sign of problems \\
        B: painful conditions will interfere with a person's ability to address problems quickly and directly \\
        C: people who have highly stressful lives tend to address problems quickly and directly \\
        \textbf{D:} refusing to think about something troubling contributes to a person's level of stress \\
        \bottomrule 
    \end{tabular}
    \caption{
        \label{tab:reclor_necessary_assumption}
        Question from ReClor with logical reasoning type: Necessary Assumption. Correct answer option in bold.
    }
\end{table}

\begin{table}
    \small
    \centering
    \begin{tabular}{p{.95\columnwidth}}
        \toprule
        \textbf{Reasoning Type: Sufficient Assumptions} \\
        \midrule
        \textbf{Context: } In Europe, schoolchildren devote time during each school day to calisthenics. North American schools rarely offer a daily calisthenics program. Tests prove that North American children are weaker, slower, and shorter-winded than European children. We must conclude that North American children can be made physically fit only if they participate in school calisthenics on a daily basis. \\
        \textbf{Question: } Which one of the following is assumed in the passage? \\
        \textbf{Options: } \\
        \textbf{A:} School calisthenics are an indispensable factor in European children's superior physical fitness. \\
        B: All children can be made physically fit by daily calisthenics. \\
        C: Superior physical fitness produces superior health. \\
        D: North American children can learn to eat a more nutritious diet as well as to exercise daily. \\
        \bottomrule 
    \end{tabular}
    \caption{
        \label{tab:reclor_sufficient_assumption}
        Question from ReClor with logical reasoning type: Sufficient Assumption. Correct answer option in bold.
    }
\end{table}

\begin{table}[h]
    \small
    \centering
    \begin{tabular}{p{.95\columnwidth}}
        \toprule
        \textbf{Reasoning Type: Strengthen} \\
        \midrule
        \textbf{Context: } Skeletal remains of early humans indicate clearly that our ancestors had fewer dental problems than we have. So, most likely, the diet of early humans was very different from ours. \\
        \textbf{Question: } Which one of the following, if true, most strengthens the argument? \\
        \textbf{Options: } \\
        A: Skeletal remains indicate that some early humans had a significant number of cavities. \\
        B: A healthy diet leads to healthy teeth. \\
        \textbf{C:} Diet is by far the most significant factor contributing to dental health. \\
        D: Early humans had a shorter average life span than we do, and the most serious dental problems now tend to develop late in life. \\
        \bottomrule 
    \end{tabular}
    \caption{
        \label{tab:reclor_strengthen}
        Question from ReClor with logical reasoning type: Strengthen. Correct answer option in bold.
    }
\end{table}

\begin{table}[h]
    \small
    \centering
    \begin{tabular}{p{.95\columnwidth}}
        \toprule
        \textbf{Reasoning Type: Weaken} \\
        \midrule
        \textbf{Context: } Many people suffer an allergic reaction to sulfites, including those that are commonly added to wine as preservatives. However, since there are several winemakers producing wine to which no sulfites are added, those who would like to drink wine but are allergic to sulfites can drink these wines without risking an allergic reaction to sulfites. \\
        \textbf{Question: } Which of the following, if true, most seriously weakens the argument? \\
        \textbf{Options: } \\
        \textbf{A:} Sulfites occur naturally in most wine. \\
        B: The sulfites that can produce an allergic reaction are also commonly found in beverages other than wine. \\
        C: Wine without added sulfites sometimes becomes undrinkable even before the wine is sold to consumers. \\
        D: Apart from sulfites, there are other substances commonly present in wine that can trigger allergic reactions. \\
        \bottomrule 
    \end{tabular}
    \caption{
        \label{tab:reclor_weaken}
        Question from ReClor with logical reasoning type: Weaken. Correct answer option in bold.
    }
\end{table}

\section{Guideline for Logical Relation Annotation}
\label{appx:map_binary_operators}
Tables~\ref{tab:pdtb_implication}, \ref{tab:pdtb_and}, and \ref{tab:pdtb_or} show mappings from natural language patterns to binary logical operators referring to the PDTB3 senses~\cite{webber2019penn}.
For the logical implication, the order of arguments is provided.
The three tables are provided to the annotators as their references during annotation. The final annotation is subject to human understanding.

\section{Indicators of Unary Operators}
\label{sec:appendix_indicators}
Table~\ref{tab:indicators} demonstrates the indicators for extracting the unary operators.

\section{Annotation Interface}
\label{appx:annotation_details}
\begin{figure*}[h]
    \centering
    \includegraphics[width=15cm]{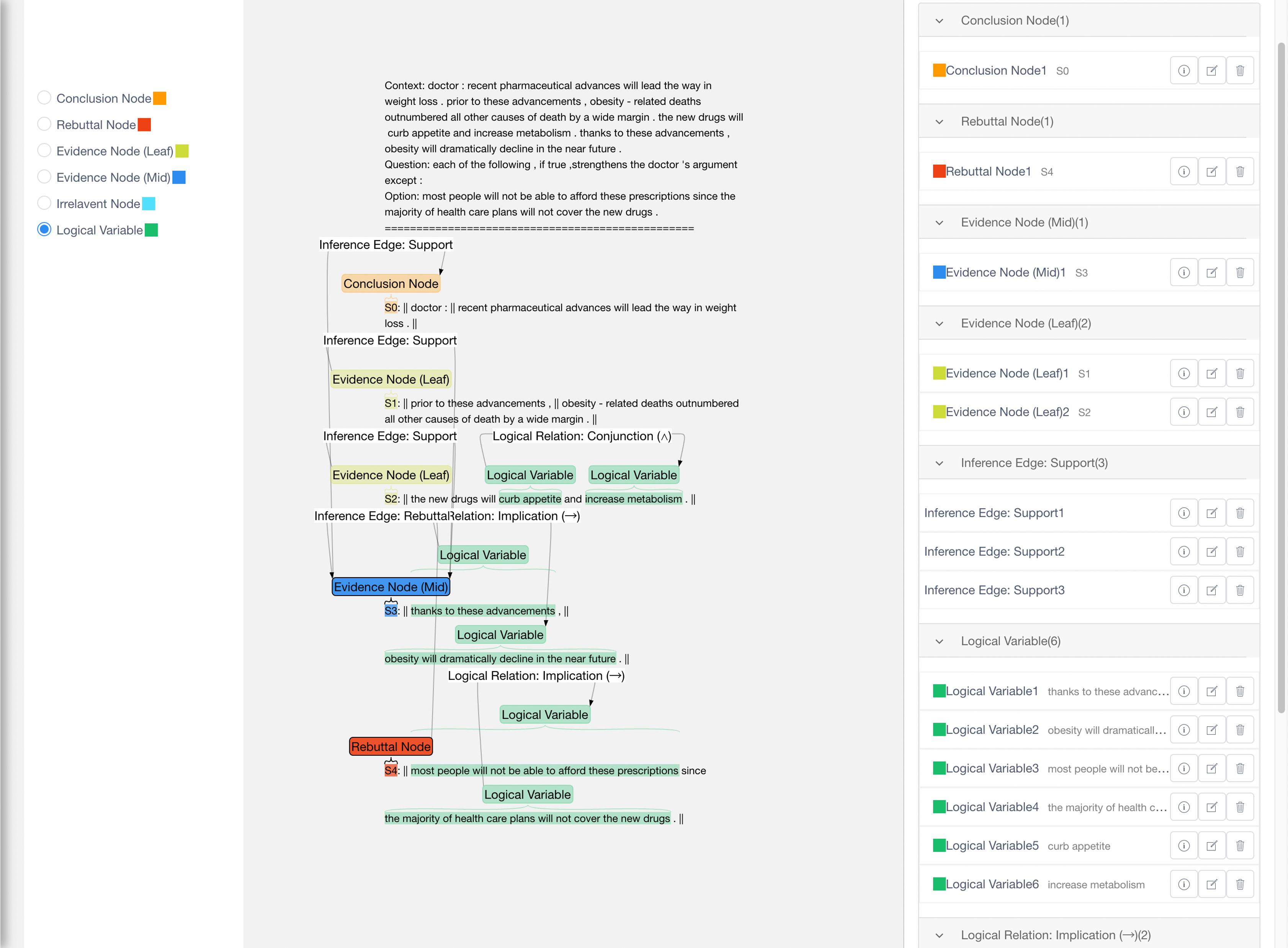}
    \caption{The annotation interface.}
    \label{fig:anno_interface}
\end{figure*}

The annotation interface is demonstrated in Figure~\ref{fig:anno_interface}.

\normalsize

\section{Model Details}
\label{appx:model_details}

\subsection{All-at-Once T5}
The input of the All-at-Once T5 is the overall logical passages with sentences and variable denotations.
The output is the linearized metagraph as shown in Table~\ref{tab:task}.
The linearized metagraph consists of three parts: the meta structure, node formula, and sentence certainty (denoted with \$graph\$, \$formula\$, and \$degree\$, respectively).
For the meta structure, we use the semicolon to connect different edges.
For the node formula, we map the operator to word using the mapping 
\{ $\Box$: [necessary],
$\Diamond$: [possible],
$\neg$: [negative],
$\land$: [and],
$\lor$: [or],
$\to$: [entail]
\}.
We use the semicolon to connect the triples for the same sentence.
We connect the certainties and formulae of sentences with ``|''.
We train the All-at-Once T5 model with a batch size of 32 and a learning rate of 1e-5 for 300 epochs.

\begin{table}[t]
  \small
  \centering
  \begin{tabular}{
    p{0.9\columnwidth}
  }
    \toprule
    \textbf{Input:} \textcolor{cyan}{{sent1}:} \textcolor{purple}{v1:} measurements of the motion of the planet uranus seem to show uranus being tugged by a force pulling it away from the sun and the inner planets . \textcolor{cyan}{{sent2}:} \textcolor{purple}{v1:} neptune and pluto , \textcolor{purple}{v2:} the two known planets whose orbits are farther from the sun than is the orbit of uranus , \textcolor{purple}{v3:} do not have enough mass to exert the force that the measurements indicate . \textcolor{cyan}{{sent3}:} \textcolor{purple}{v1:} therefore , \textcolor{purple}{v2:} in addition to the known planets , \textcolor{purple}{v3:} there must be at least one planet in our solar system that we have yet to discover . \textcolor{cyan}{{sent4}:} \textcolor{purple}{v1:} there is a belt of comets beyond the orbit of pluto with powerful gravitational pull . \\
    \midrule
    \textbf{Output:} \textcolor{cyan}{\$graph\$} sent1 -> sent3; sent2 -> sent3; sent4 => sent2;  \textcolor{purple}{\$formula\$} sent3: v2 [and] [necessary] v3;  \textcolor{olive}{\$degree\$} sent1: contingent | sent2: contingent | sent3: necessary | sent4: contingent \\
    \bottomrule
  \end{tabular}
  \caption{
    \label{tab:task}
    Example of the metagraph generation task. 
    ``->'' denotes the \textit{support} relation, and ``=>'' denotes the \textit{rebut} relation.
    The readable triple in \texttt{sent3} formulas is $\textrm{v}_2\land\Box \textrm{v}_3$.
  }
\end{table}

\subsection{Multitask T5}
The Multitask T5 decomposes the whole generation task into three sub-tasks: meta structure generation, formula generation, and certainty prediction.
We train a single T5 model on these three sub-tasks simultaneously.
We follow~\citet{raffel2020exploring} to add a task-specific prefix to the input before feeding it to the model.
Table~\ref{tab:multitask_example} shows some specific input and output examples of each sub-task.
We train the Multitask T5 model with a batch size of 32 and a learning rate of 1e-5 for 300 epochs.
We use the examples-proportional mixing~\cite{raffel2020exploring} and simply concatenate the data for all sub-tasks as the training data for Multitask T5.

\begin{table}[t]
\small
\centering
\begin{tabular}{p{.9\columnwidth}}
    \toprule
    \textbf{Task:} \textbf{Meta structure generation }\\
    \textbf{Input:} \texttt{GRAPH:} sent1: to reduce waste of raw materials , the government of sperland is considering requiring household appliances to be broken down for salvage when discarded . [AND] imposing the fee at the time of salvage would reduce waste more effectively , however , because consumers tend to keep old appliances longer if they are faced with a fee for discarding them .  sent2: to cover the cost of salvage , the government is planning to charge a fee , which would be imposed when the appliance is first sold . sent4: increasing the cost of disposing of an appliance properly increases the incentive to dispose of it improperly . \\
    \textbf{Output:} sent4 => sent1; sent1 -> sent2; \\ 
    
    \midrule
    \textbf{Task:} \textbf{Formula generation} \\
    \textbf{Input:} \texttt{FORMULAE:} v1: grammarians have for years condemned as ungrammatical the english phrase " between you and i " , insisting that the correct phrasing is " between you and me , " with v2: the objective case after v3: a preposition . v4: such condemnations , however , are obviously unfounded , because v5: shakespeare himself , in the merchant of venice , wrote , " all debts are cleared between you and i. "\\
    \textbf{Output:} [necessary] v5 [entail] [necessary] v4;  v2 [entail]  v3; \\ 
    
    \midrule
    \textbf{Task:} \textbf{Certainty prediction} \\
    \textbf{Input:} \texttt{DEGREE:} it was formerly believed that prehistoric homo sapiens ancestors of contemporary humans interbred with neanderthals , but dna testing of a neanderthal ' s remains indicates that this is not the case . \\
    \textbf{Output:} impossible \\ 
    \bottomrule 
\end{tabular}
\caption{
Input and output examples of all the sub-tasks of Multitask T5.
}
\label{tab:multitask_example}
\end{table}

\subsection{MetGen}
MetGen~\cite{hong2022metgen} is a module-based framework for structured explanation generation.

\noindent\textbf{Modules.}
We use two types of modules: the conclusion module and the rebuttal module.
The conclusion module takes two sentences as input (e.g., \texttt{sent1:\dots \ sent2:\dots}) 
and outputs the inference relation type between them (e.g., \texttt{sent1 -> sent2} or \texttt{sent2 -> sent1}).
If there is no conclusive relationship between the two input sentences, the module would output the word \texttt{none}.
The rebuttal module is defined similarly.

\noindent\textbf{Controller.}
The controller decides the reasoning direction based on the current reasoning state.
Specifically, given the current partially metagraph and all the sentences, the controller predicts which combinations of two sentences (e.g., \texttt{sent1 sent2}) should be considered in the next step.
If the current proof is complete, the controller would output the word \texttt{done}.

\noindent\textbf{Reasoning Process.}
MetGen generates the metagraphs in an iterative manner.
It iteratively repeats the following reasoning iteration to grow the graph until either the controller returns \texttt{done} or the maximum number of iteration steps is reached.
It takes several iterations before completing the generation.
In each iteration, MetGen generates one step.
It first uses the controller to predict some possible sentence combinations.
Then, each combination is sent to the reasoning modules to generate candidate steps that indicate the detailed inference relation type between them.
The candidate step with the highest score (the lowest perplexity) is picked for the next iteration.

\noindent\textbf{Implementations.}
To compare with other methods under the same number of parameters, we implement MetGen using a single T5 model.
Table~\ref{tab:metgen_example} shows some input and output examples of MetGen.
The MetGen is trained on five sub-tasks simultaneously: controller task, conclusion module task, rebuttal module task, formula generation task, and certainty prediction tasks.
We train the MetGen model with a batch size of 32 and a learning rate of 1e-5 for 300 epochs.
We set the maximum number of iteration steps as 3.

\begin{table}[h]
\small
\centering
\begin{tabular}{p{.9\columnwidth}}
    \toprule
    \textbf{Task:} \textbf{Controller}\\
    \textbf{Input:} \texttt{CONTROL:} proof: sent1 -> sent3; context: sent1: measurements of the motion of the planet uranus seem to show uranus being tugged by a force pulling it away from the sun and the inner planets . sent2: neptune and pluto , the two known planets whose orbits are farther from the sun than is the orbit of uranus , do not have enough mass to exert the force that the measurements indicate . sent3: therefore , in addition to the known planets , there must be at least one planet in our solar system that we have yet to discover . sent4: there is a belt of comets beyond the orbit of pluto with powerful gravitational pull . \\
    \textbf{Output:} sent2 sent3 \\
    
    \midrule
    \textbf{Task:} \textbf{Conclusion Module} \\
    \textbf{Input:} \texttt{CONCLUSION:} sent2: the dna of contemporary humans is significantly different from that of the neanderthal. sent3: the dna of prehistoric homo sapiens ancestors of contemporary humans was not significantly more similar to that of neanderthals than is the dna of contemporary humans. \\
    \textbf{Output:} sent3 -> sent2 \\ 
    
    \midrule
    \textbf{Task:} \textbf{Rebuttal Module} \\
    \textbf{Input:} \texttt{REBUTTAL:} sent1: recent unexpectedly heavy rainfalls in the metropolitan area have filled the reservoirs and streams ; water rationing , therefore , will not be necessary this summer . sent2: the water company 's capacity to pump water to customers has not kept up with the increased demand created by population growth in the metropolitan area . \\
    \textbf{Output:} sent2 => sent1 \\ 
    \bottomrule 
\end{tabular}
\caption{
Input and output examples of the controller and modules of MetGen.
MetGen is also trained with the formula generation and certainty prediction tasks.
}
\label{tab:metgen_example}
\end{table}

\subsection{Experimental Details}
We use the pre-trained models from \texttt{HuggingFace Transformers}\footnote{https://github.com/huggingface/transformers}.
We use the Adafactor optimizer~\cite{shazeer2018adafactor}.
We run the experiments based on T5-large 3 times with different random seeds and report the average performances.
The experiments based on T5-11b are run only once considering the computational cost.

\section{Evaluation Metrics}
\label{sec:evaluation_metrics}

\noindent\textbf{Meta structure:}
Does the predicted metagraph use the correct sentences and have the correct structure?
For meta nodes, we report a node F1 score by comparing the set of sentences used in the predicted and gold metagraph.
For meta structure, we decompose the metagraph into one-premise steps (e.g., \texttt{sent1 -> sent2}).
We compare the set of steps in the predicted and gold metagraph and report the step F1 score.
A predicted step is correct if its premise, conclusion, and step type match the gold one.
The AllCorrect score is 1 if the F1 is 1, 0 otherwise.

\noindent\textbf{Formula:}
Does the predicted metagraph have the correct internal structure of meta nodes?
For each sentence, we measure the formula F1 score by comparing all formulae in the predictions and gold annotations.
A predicted formula is considered correct if its certainty operators, binary operators, and variables match the gold one.
For the certainty operators, we reduce them to the standard form (one of the five degree-of-certainty forms listed in Table~\ref{tab:modal}) before comparison.
For the binary operator, we consider its symmetry.
For example, $\neg \textrm{v}_1 \land \Diamond \textrm{v}_2$ is equivalent to $\Diamond \textrm{v}_2 \land \neg \textrm{v}_1$, but $\neg \textrm{v}_1 \to \Diamond \textrm{v}_2$ is not equivalent to $\Diamond \textrm{v}_2 \to \neg \textrm{v}_1$.
The AllCorrect score is 1 if the formula F1 is 1, 0 otherwise.
Since each sample contains multiple sentences, we average the formula F1 scores of all sentences in the sample as the formula F1 score for this sample.

\noindent\textbf{Certainty:}
Are the certainties of the sentence correct?
For each sample, we compute the accuracy of the predicted certainties of the sentences.
The AllCorrect score is 1 if the accuracy is 1, and 0 otherwise.
We report the accuracy and AllCorrect score of the testing dataset, which is the average accuracy and AllCorrect score of all samples in the dataset.
Due to the unbalance of the certainty labels, we gather the predictions for all sentences in the dataset (ignoring which sample the sentence comes from) and report the macro-F1 score.

\noindent\textbf{Overall:}
The overall AllCorrect score of a predicted metagraph is 1 only if all of the meta structure, formulae, and certainties are correct.
This is a strict metric since any error would result in a score of 0.

\section{Detailed Analysis Results}
\label{appx:ablation}

Table~\ref{tab:exp_steps_per_operator} present the detailed performance of different inference steps and different operators.
Table~\ref{tab:fewshot} and Table~\ref{tab:fewshot_steps} shows the detailed results with different ratios of training data.

\begin{table}[t]
    \small
	\centering
	\begin{tabular}{
		|p{0.15\columnwidth}|
		p{0.7\columnwidth}|
	}
		\hline
		\makecell[l]{Negation \\($\neg$)} & \texttt{"no", "not", "none", "nobody", "nothing", "neither", "nor", "nowhere", "never", "hardly", "scarcely", "barely", "doesn't", "isn't", "wasn't", "shouldn't", "wouldn't", "couldn't", "won't", "can't", "don't", "impossible"} \\
		\hline
		\makecell[l]{Box \\($\Box$)} & \texttt{"necessarily", "must", "definitely", "certainly", "clearly", "obviously", "undoubtedly", "surely", "will", "all", "every", "always"} \\
		\hline
		\makecell[l]{Diamond \\($\Diamond$)} & \texttt{"likely", "approximately", "possibly", "perhaps", "probably", "maybe", "few", "may", "might", "could", "many", "most", "some", "numerous", "countless", "majority", "often", "frequently", "commonly", "usually", "sometimes", "repeatedly", "appears", "seems", "suggests", "indicates"} \\
		\hline
	\end{tabular}
	\caption{
		\label{tab:indicators}
		Indicators of unary operators. 
	}
\end{table}

\begin{table*}[!t]
    \small
    \centering
    \begin{tabular}{
        c|cccc|cccccccc
    }
         \toprule
         \multirow{2}*{} & \multicolumn{2}{c}{Support} & \multicolumn{2}{c|}{Rebut} & \multicolumn{7}{c}{Operators} \\
         & F1 & All & F1 & All & $\to$ & $\land$ & $\lor$ & $\neg$ & $\Box$ & $\Diamond$ & N/A \\ \midrule
         Once (large) & 48.9 & 15.8 & 53.3 & 53.0 & 25.4 & 23.8 & 2.1 & 3.2 & 3.4 & 7.5 & 83.1 \\
         Multitask (large) & 56.4 & 22.5 & 57.0 & 56.8 & 52.2 & 38.2 & 27.1 & 13.3 & 6.8 & 13.7 & 98.4 \\
         MetGen (large) & 60.9 & 26.8 & 59.8 & 59.3 & 58.8 & 41.9 & 33.3 & 14.3 & 11.4 & 24.9 & 96.3 \\
         \midrule
         Once (11b) & 57.1 & 27.0 & 59.5 & 59.5 & 42.6 & 34.5 & 18.8 & 3.2 & 6.3 & 13.5 & 93.6 \\
        Multitask (11b) & 62.8 & 34.0 & 64.2 & 63.5 & 52.0 & 41.9 & 31.2 & 7.1 & 9.4 & 15.3 & 99.3 \\
        MetGen (11b) & 65.6 & 32.5 & 69.7 & 69.0 & 60.8 & 52.5 & 43.8 & 12.7 & 15.6 & 18.8 & 97.2 \\
         \bottomrule
    \end{tabular}
    \caption{
        \label{tab:exp_steps_per_operator}
        Performance on two types of inference step.
        Per-operator F1 scores.
    }
\end{table*}

\begin{table*}[]
    \small
    \centering
    \begin{tabular}{
    c|c|cccc|cc|ccc|c
    }
        \toprule
       & & \multicolumn{4}{c|}{Component 1} & \multicolumn{2}{c|}{Component 2} & \multicolumn{3}{c|}{Component 3} & \multirow{2}*{Overall} \\
       & & \multicolumn{2}{c}{Node} & \multicolumn{2}{c|}{Step} & \multicolumn{2}{c|}{Formulae} & \multicolumn{3}{c|}{Certainty} & \\ 
       & Ratio & F1 & All & F1 & All & F1 & All & Acc & All & F1$^*$ & All \\ \midrule
      \multirow{6}*{Once (large)} 
       & 50\% & 89.2 & 50.5 & 39.0 & 14.0 & 52.5 & 51.6 & 73.2 & 34.0 & 22.4 & 3.6 \\
       & 20\% & 86.6 & 45.5 & 36.1 & 11.0 & 44.6 & 43.7 & 70.7 & 29.0 & 18.3 & 3.5 \\
       & 10\% & 88.8 & 49.0 & 30.1 & 6.5 & 42.0 & 41.8 & 65.6 & 25.5 & 18.2 & 1.2 \\
       & 5\% & 82.3 & 26.0 & 20.0 & 0.5 & 40.2 & 40.0 & 66.8 & 20.0 & 14.7 & 0.5 \\
       & 2\% & 71.0 & 5.5 & 14.6 & 0.0 & 31.8 & 31.4 & 65.3 & 18.0 & 13.7 & 0.0 \\
       & 1\% & 70.7 & 2.5 & 13.6 & 0.5 & 38.9 & 38.8 & 64.7 & 14.5 & 15.8 & 0.0 \\
      \midrule
      \multirow{6}*{Multitask (large)} 
       & 50\% & 94.1 & 68.5 & 55.2 & 22.0 & 78.9 & 76.9 & 84.9 & 57.5 & 50.8 & 10.5 \\
       & 20\% & 92.4 & 64.0 & 44.7 & 16.0 & 70.3 & 68.7 & 81.0 & 47.0 & 22.9 & 5.0 \\
       & 10\% & 56.1 & 3.5 & 13.3 & 1.5 & 71.1 & 70.2 & 80.3 & 47.0 & 23.0 & 0.3 \\
       & 5\% & 38.9 & 2.0 & 5.7 & 0.5 & 65.9 & 65.0 & 80.2 & 45.0 & 18.7 & 0.0 \\
       & 2\% & 31.1 & 2.5 & 3.9 & 1.0 & 61.5 & 60.6 & 80.7 & 46.0 & 18.6 & 0.3 \\
       & 1\% & 8.6 & 2.0 & 2.0 & 1.0 & 56.7 & 56.4 & 80.8 & 46.0 & 18.0 & 0.0 \\
      \midrule
      \multirow{6}*{MetGen (large)} 
       & 50\% & 94.3 & 67.0 & 53.1 & 17.5 & 75.5 & 74.1 & 83.5 & 56.5 & 31.1 & 9.2 \\
       & 20\% & 92.4 & 64.0 & 44.7 & 16.0 & 70.3 & 68.7 & 81.0 & 47.0 & 22.9 & 5.0 \\
       & 10\% & 56.1 & 3.5 & 13.3 & 1.5 & 71.1 & 70.2 & 80.3 & 47.0 & 23.0 & 0.3 \\
       & 5\% & 38.9	& 2.0 & 5.7 & 0.5 & 65.9 & 65.0 & 80.2 & 45.0 & 18.7 & 0.0 \\
       & 2\% & 31.1	& 2.5 & 3.9 & 1.0 & 61.5 & 60.6 & 80.7 & 46.0 & 18.6 & 0.3 \\
       & 1\% & 8.6 & 2.0 & 2.0 & 1.0 & 56.7 & 56.4 & 80.8 & 46.0 & 18.0 & 0.0 \\
      \bottomrule
    \end{tabular}
    \caption{
        \label{tab:fewshot}
        Detailed results on different ratios of \ourdataset training data.
        $^*$: macro-F1. 
        }
\end{table*}

\begin{table*}[!t]
    \small
    \centering
    \begin{tabular}{
        p{0.15\columnwidth}
        |c|cccc|cccccccc
    }
         \toprule
         & \multirow{2}*{} & \multicolumn{2}{c}{Support} & \multicolumn{2}{c|}{Rebut} & \multicolumn{7}{c}{Operators} \\
         & Ratio & F1 & All & F1 & All & $\to$ & $\land$ & $\lor$ & $\neg$ & $\Box$ & $\Diamond$ & N/A \\ 
         \midrule
         \multirow{6}*{\makecell[c]{Once\\(large)}}
            & 50\% & 42.2 & 16.0 & 47.8 & 47.5 & 19.1 & 13.8 & 0.0 & 0.0 & 0.0 & 2.8 & 79.4 \\
            & 20\% & 38.4 & 12.0 & 49.3 & 49.0 & 11.3 & 6.7 & 0.0 & 3.2 & 2.1 & 6.9 & 71.6 \\
            & 10\% & 31.7 & 7.5 & 50.3 & 50.0 & 5.9 & 5.6 & 0.0 & 9.5 & 2.1 & 0.0 & 69.7 \\
            & 5\% & 20.4 & 1.5 & 26.3 & 26.0 & 5.5 & 1.8 & 0.0 & 0.0 & 0.0 & 0.0 & 68.8\\
            & 2\% & 17.9 & 1.0 & 15.5 & 15.5 & 3.7 & 5.7 & 0.0 & 0.0 & 0.0 & 0.0 & 53.4 \\
            & 1\% & 17.2 & 1.5 & 12.5 & 12.5 & 4.3 & 1.4 & 0.0 & 0.0 & 2.1 & 2.1 & 65.7 \\
         \midrule
         \multirow{6}*{\makecell[c]{Multitask\\(large)}}
            & 50\% & 59.2 & 24.5 & 60.3 & 60.0 & 57.2 & 47.4 & 37.5 & 15.1 & 17.2 & 26.0 & 97.2 \\
            & 20\% & 48.9 & 18.5 & 52.5 & 52.5 & 36.2 & 22.3 & 18.8 & 0.0 & 0.0 & 3.8 & 98.3 \\
            & 10\% & 16.9 & 6.5 & 56.0 & 56.0 & 36.1 & 22.3 & 6.2 & 4.8 & 0.0 & 2.1 & 99.5 \\
            & 5\% & 7.6 & 3.0 & 58.0 & 58.0 & 30.2 & 14.6 & 12.5 & 0.0 & 1.6 & 8.3 & 96.2 \\
            & 2\% & 4.8 & 2.5 & 58.0 & 58.0 & 34.5 & 12.7 & 0.0 & 0.0 & 3.6 & 4.9 & 87.5 \\
            & 1\% & 4.2	& 3.5 & 59.0 & 59.0 & 25.6 & 5.4 & 0.0 & 0.0 & 2.1 & 2.8 & 85.3 \\
         \midrule
         \multirow{6}*{\makecell[c]{MetGen\\(large)}}
            & 50\% & 58.1 & 20.5 & 58.5 & 58.0 & 47.8 & 34.5 & 25.0 & 12.7 & 1.6 & 11.1 & 98.6 \\
            & 20\% & 53.3 & 20.5 & 56.2 & 55.5 & 43.1 & 28.5 & 18.8 & 0.0 & 3.1 & 7.9 & 97.6 \\
            & 10\% & 55.3 & 20.5 & 55.8 & 55.5 & 35.5 & 26.2 & 6.2 & 4.8 & 0.0 & 4.2 & 99.3 \\
            & 5\% & 50.5 & 16.0 & 55.8 & 55.5 & 32.1 & 23.0 & 6.2 & 0.0 & 0.0 & 3.5 & 93.9 \\
            & 2\% & 40.2 & 3.5 & 58.5 & 58.5 & 28.3 & 13.6 & 0.0 & 0.0 & 1.6 & 6.2 & 91.0 \\
            & 1\% & 35.0 & 3.0 & 59.0 & 59.0 & 23.4 & 3.0 & 0.0 & 0.0 & 2.1 & 2.8 & 89.1 \\
         \bottomrule
    \end{tabular}
    \caption{
        \label{tab:fewshot_steps}
        Inference step and operator performances on different ratios of \ourdataset training data.
    }
\end{table*}

\section{Error Cases}
\label{appx:error_case}

Examples of each error type are shown in Tables~\ref{tab:error_case_graph},~\ref{tab:error_case_formulae}, and~\ref{tab:error_case_degree}.

\begin{table*}[!t]
    \small
    \centering
    \begin{tabular}{
        p{.9\textwidth}
    }
        \toprule
         \midrule
        \textbf{G1: Incorrect Inference Type }\\ \midrule
         \textbf{Passage:} sent1: for similar cars and drivers , automobile insurance for collision damage has always cost more in greatport than in fairmont . [AND] police studies , however , show that cars owned by greatport residents are , on average , slightly less likely to be involved in a collision than cars in fairmont .  sent3: clearly , therefore , insurance companies are making a greater profit on collision - damage insurance in greatport than in fairmont . sent4: repairing typical collision damage does not cost more in greatport than in fairmont . \\ 
         \textbf{Gold:} sent4 -> sent3; sent1 -> sent3; \\ 
         \textbf{Pred:} sent1 -> sent3; sent4 => sent3; \\ \midrule
        \midrule
        \textbf{G2: Incorrect Rebuttal} \\ \midrule
         \textbf{Passage:} sent1: there should be a greater use of gasohol . sent2: gasohol is a mixture of alcohol and gasoline , and has a higher octane rating and fewer carbon monoxide emissions than straight gasoline . [AND] burning gasohol adds no more carbon dioxide to the atmosphere than plants remove by photosynthesis .  sent4: cars burn on the average slightly more gasohol per kilometer than they do gasoline .\\ 
         \textbf{Gold:} sent2 -> sent1; sent4 => sent2; \\
         \textbf{Pred:} sent4 -> sent1; sent2 -> sent1; \\ \midrule
        \midrule
        \textbf{G3: Incorrect Conclusion} \\ \midrule
         \textbf{Passage:} sent1: healthy lungs produce a natural antibiotic that protects them from infection by routinely killing harmful bacteria on airway surfaces . [AND] people with cystic fibroses , however , are unable to fight off such bacteria , even though their lungs produce normal amounts of the antibiotic .  sent3: since the fluid on airway surfaces in the lungs of people with cystic fibrosis has an abnormally high salt concentration , scientists hypothesize that in high salt environments the antibiotic becomes ineffective at killing harmful bacteria . sent4: the lungs of people who suffer from cystic fibrosis are unable to fight off harmful bacteria even when the salt concentration is reduced to levels typical of healthy lungs . \\
         \textbf{Gold:} sent3 -> sent1; sent4 => sent3; \\
         \textbf{Pred:} sent4 -> sent3; sent1 -> sent3; \\ \midrule
         \midrule
        \textbf{G4: Incorrect Inference Step} \\ \midrule
         \textbf{Passage:} sent1: spokesperson : the major school lunch vendors recently agreed to stop selling high - calorie beverages in elementary and middle schools because studies show that children of ages 7 to 8 who substitute one low - calorie beverage for one high - calorie soft drink in their daily diets will , on average , weigh 20 pounds less than they would have by the time they reach high school . sent2: since only low - calorie beverages will be sold in schools , within six to eight years , we can expect to see a reduction in the percentage of overweight high - school children . sent3: elementary and middle school students who used to buy high - calorie soft drinks at school will not bring them to school or drink extra high - calorie beverages at home as a substitute . \\
         \textbf{Gold:} sent3 -> sent2; sent1 -> sent2; \\
         \textbf{Pred:} sent3 -> sent2; \\ \midrule
         \midrule
       \textbf{G5: Other Structural Mismatch} \\ \midrule
        \textbf{Passage:} sent1: employer : in the current economic climate , the best way to run a business is to pay employees the least amount possible to do the job . sent2: the supply of labor is far outpacing demand since the number of college graduates increases every year and the average age of retirement is also increasing . [AND] applicants will typically take the first job offer on the table , and any employee who demands a raise can be easily replaced from the labor pool .  sent4: even if the employee is unhappy , he or she will often remain on the job due to the competition in the job market . [AND] keeping payroll costs low allows more resources to be devoted to innovation , delivering a higher quality product to customers .  sent6: automation is the leading cause for unemployment . \\
        \textbf{Gold:} sent4 -> sent1; sent2 -> sent4; \\
        \textbf{Pred:} sent6 => sent1; sent2 -> sent1; \\ \midrule
         \bottomrule
    \end{tabular}
    \caption{Error cases for meta graph structure.}
    \label{tab:error_case_graph}
\end{table*}

\begin{table*}[!t]
    \small
    \centering
    \begin{tabular}{
        p{.9\textwidth}
    }
        \toprule
         \midrule
        \textbf{F1: Incorrect Logical Variable} \\ \midrule
         \textbf{Sentence:} v1: legislators considering a proposed law for which they have v2: repugnance or v3: enthusiasm v4: do not consider the consequences that it will actually have . \\ 
         \textbf{Gold:} v3 [or]  v2; \\ 
         \textbf{Pred:} v2 [or] v3; v4 [entail] v1; \\ \midrule
        \midrule
        \textbf{F2: Incorrect Unary Operator} \\ \midrule
         \textbf{Sentence:} v1: auditor : xyz , v2: a construction company , purchased 20 new trucks 3 years ago , and v3: there is no record of any of those trucks being sold last year . \\ 
         \textbf{Gold:} v2 [and]  v3; \\
         \textbf{Pred:} v1 [and] v3; \\ \midrule
        \midrule
        \textbf{F3: Incorrect Binary Operator} \\ \midrule
         \textbf{Sentence:} v1: travaillier corporation has recently hired employees with experience in the bus tour industry v2: its executives have also been negotiating with charter bus companies that subcontract with bus tour companies . [AND] but v3: travaillier has traditionally focused on serving consumers who travel primarily by air , and v4: marketing surveys show that travaillier ' s traditional consumers have not changed their vacation preferences . \\
         \textbf{Gold:} v1 [and]  v2;  v3 [and]  v4; \\
         \textbf{Pred:} v3 [and] v4; v2 [entail] v1; \\ \midrule
         \midrule
        \textbf{F4: Incorrect Implication Direction} \\ \midrule
         \textbf{Sentence:} v1: now some politicians are saying that , in order to v2: cause another similarly sized increase in exports , v3: the government should allow the pundra to become weak again . \\
         \textbf{Gold:} v3 [entail]  v2; \\
         \textbf{Pred:} v2 [entail] v3; \\
         \midrule
         \bottomrule
    \end{tabular}
    \caption{Error cases for the formulae.}
    \label{tab:error_case_formulae}
\end{table*}

\begin{table*}[!t]
    \small
    \centering
    \begin{tabular}{
        p{.9\textwidth}
    }
        \toprule
         \midrule
        \textbf{C1: Incorrect Polarity} \\ \midrule
         \textbf{Sentence:} the chemistry department 's funding for basic science research is not likely to increase if its funding from sources other than profit - driven institutions does not increase . \\ 
         \textbf{Gold:} impossible \\
         \textbf{Pred:} possible \\ \midrule
        \midrule
        \textbf{C2: Other Polarities to Contingent} \\ \midrule
         \textbf{Sentence:} if legislators are to enact laws that benefit constituents , they must be sure to consider what the consequences of enacting a proposed law will actually be . [AND] concerned primarily with advancing their own political careers , legislators present legislation in polemical terms ; this arouses in their colleagues either repugnance or enthusiasm for the legislation . \\
         \textbf{Gold:} necessary \\
         \textbf{Pred:} contingent \\ \midrule
        \midrule
        \textbf{C3: Contingent to Other Polarities} \\ \midrule
         \textbf{Sentence:} making decisions about patterns of work organization , resource allocation , and location of industry is not the core of a public official 's job . \\
         \textbf{Gold:} contingent \\
         \textbf{Pred:} unnecessary \\ \midrule
         \midrule
        \textbf{C4: Unresolved Certainty} \\ \midrule
         \textbf{Sentence:} the link between jogging and certain structural disorders appears to be a causal one . \\
         \textbf{Gold:} contingent \\
         \textbf{Pred:} causal \\ 
         \midrule
         \bottomrule
    \end{tabular}
    \caption{Error cases for certainty.}
    \label{tab:error_case_degree}
\end{table*}

\onecolumn

\small
\onecolumn
\begin{longtable}{  
    lcl
  }
  \toprule
  \textbf{Connectives} & \textbf{Binary Operator} & \textbf{Senses} \\
  \midrule
  about & $A\to B$ & Contingency.Cause.Reason \\ \hline
  \makecell[l]{A accordingly B. \\A; accordingly B. \\A. B accordingly.} & $A\to B$ & Contingency.Cause.Result \\ \hline
  after & $A\to B$ & \makecell[l]{Temporal.Asynchronous.Succession \\Contingency.Cause.Reason} \\ \hline
  afterward & $A\to B$ & Temporal.Asynchronous.Precedence \\ \hline
  afterwards & $A\to B$ & Temporal.Asynchronous.Precedence \\ \hline
  \makecell[l]{B, as A. \\As A, B.} & $A\to B$ & \makecell[l]{Contingency.Cause+Belief.Reason+Belief \\Contingency.Cause.Reason \\Expansion.Instantiation.Arg2-as-instance \\Expansion.Level-of-detail.Arg1-as-detail \\Expansion.Manner.Arg2-asmanner \\Temporal.Asynchronous.Succession \\Contingency.Cause.Reason \\Temporal.Asynchronous.Succession} \\ \hline
  as a result & $A\to B$ & Contingency.Cause.Result \\ \hline
  \makecell[l]{B, as if A. \\As if A, B.} & $A\to B$ & \makecell[l]{Comparison.Concession.Arg2-as-denier \\Comparison.Similarity \\Expansion.Manner.Arg2-asmanner \\Expansion.Instantiation.Arg1-as-instance \\Expansion.Manner.Arg2-asmanner} \\ \hline
  \makecell[l]{Because of A, B. \\B because of A.} & $A\to B$ & Contingency.Cause.Reason \\ \hline
  \makecell[l]{Because A, B. \\B because A.} & $A\to B$ & \makecell[l]{Contingency.Cause+Belief.Reason+Belief \\Contingency.Cause.Reason \\Contingency.Condition+SpeechAct} \\ \hline
  \makecell[l]{B, as long as A. \\As long as A, B.} & $A\to B$ & Contingency.Condition.Arg2 -as-cond \\ \hline
  \makecell[l]{Before B, A. \\A before B.} & $A\to B$ & \makecell[l]{Temporal.Asynchronous.Precedence \\Temporal.Asynchronous.Succession} \\ \hline
  \makecell[l]{By A, B. \\B by A.} & $A\to B$ & \makecell[l]{Contingency.Cause+Belief.Reason+Belief \\Expansion.Manner.Arg2-asmanner \\Contingency.Cause.Reason \\Expansion.Manner.Arg2-asmanner \\Contingency.Cause.Reason \\Contingency.Condition.Arg2-as-cond \\Expansion.Manner.Arg2-asmanner \\Contingency.Condition.Arg2-as-cond \\Contingency.Purpose.Arg1as-goal \\Expansion.Manner.Arg2-asmanner \\Expansion.Level-of-detail.Arg2-as-detail \\Expansion.Manner.Arg2-asmanner} \\ \hline
  by then & $A\to B$ & \makecell[l]{Temporal.Asynchronous.Succession \\Contingency.Cause.Reason \\Temporal.Asynchronous.Succession} \\ \hline
  A consequently B. & $A\to B$ & Contingency.Cause.Result \\ \hline
  \makecell[l]{B depending on A. \\Depending on A, B.} & $A\to B$ & Contingency.Condition.Arg2-as-cond \\ \hline
  \makecell[l]{B depending upon A. \\Depending upon A, B.} & $A\to B$ & Contingency.Condition.Arg2-as-cond \\ \hline

  \makecell[l]{B, due to A. \\Due to A, B.} & $A\to B$ & Contingency.Cause.Reason \\ \hline
  \makecell[l]{B. Earlier, A. \\B, A earlier.} & $A\to B$ & Temporal.Asynchronous.Su ccession \\ \hline
  \makecell[l]{B even after A. \\Even after A, B.} & $A\to B$ & \makecell[l]{Temporal.Asynchronous.Succession \\Comparison.Concession.Arg1-as-denier} \\ \hline
  A, finally, B. & $A\to B$ & \makecell[l]{Temporal.Asynchronous.Precedence \\Contingency.Cause.Result \\Temporal.Asynchronous.Precedence} \\ \hline
  \makecell[l]{B, for A. \\For A, B.} & $A\to B$ & \makecell[l]{Comparison.Concession.Arg1-as-denier \\Contingency.Cause.Reason Contingency.Cause.Result \\Contingency.Condition.Arg2 -as-cond \\Contingency.Purpose.Arg2as-goal \\Expansion.Level-of-detail.Arg2-as-detail} \\ \hline
  B, for example A. & $A\to B$ & Expansion.Instantiation.Arg2-as-instance \\ \hline
  B, for instance A. & $A\to B$ & Expansion.Instantiation.Arg 2-as-instance \\ \hline
  B, from A. & $A\to B$ & \makecell[l]{Contingency.Cause+Belief.Reason+Belief \\Contingency.Cause.Reason \\Contingency.Condition.Arg2-as-cond \\Contingency.Cause.Reason \\Expansion.Manner.Arg2-asmanner \\Contingency.Cause.Reason \\Expansion.Substitution.Arg1-as-subst} \\ \hline
  \makecell[l]{Given A, B. \\B, given A.} & $A\to B$ & \makecell[l]{Contingency.Cause+Belief.Reason+Belief \\Contingency.Cause.Reason} \\ \hline
  A, hence B. & $A\to B$ & Contingency.Cause.Result \\ \hline
  \makecell[l]{If A, B. \\B, if A.} & $A\to B$ & \makecell[l]{Comparison.Concession+SpeechAct.\\\quad Arg2-as-denier+SpeechAct \\Comparison.Concession.Arg1-as-denier \\Comparison.Concession.Arg2-as-denier \\Comparison.Contrast \\Contingency.Condition+SpeechAct \\Contingency.Condition.Arg2-as-cond \\Expansion.Level-of-detail.Arg2-as-detail \\Contingency.Condition.Arg2-as-cond \\Temporal.Synchronous \\Contingency.Condition.Arg2-as-cond} \\ \hline
  B if and when A. & $A\to B$ & \makecell[l]{Contingency.Condition.Arg2-as-cond \\Temporal.Synchronous \\Contingency.Condition.Arg2-as-cond} \\ \hline
  B if and when A. & $A\to B$ & \makecell[l]{Contingency.Condition.Arg2-as-cond \\Temporal.Synchronous \\Contingency.Condition.Arg2-as-cond} \\ \hline
  B if only A. & $A\to B$ & \makecell[l]{Comparison.Concession.Arg2-as-denier \\Contingency.Condition.Arg2-as-cond \\Contingency.Purpose.Arg2as-goal} \\ \hline
  If A then B. & $A\to B$ & \makecell[l]{Contingency.Condition+SpeechAct \\Contingency.Condition.Arg2-as-cond} \\ \hline
  B in A. & $A\to B$ & \makecell[l]{Contingency.Cause+Belief.Reason+Belief \\Contingency.Cause.Reason \\Expansion.Manner.Arg2-asmanner \\Contingency.Cause.Reason \\Contingency.Condition.Arg2-as-cond \\Expansion.Manner.Arg2-asmanner \\Contingency.Condition.Arg2-as-cond \\Contingency.Purpose.Arg2as-goal \\Expansion.Instantiation.Arg1-as-instance \\Expansion.Level-of-detail.Arg1-as-detail \\Expansion.Level-of-detail.Arg2-as-detail \\Expansion.Manner.Arg2-asmanner \\Temporal.Synchronous \\Contingency.Purpose.Arg2as-goal \\Temporal.Synchronous \\Expansion.Level-of-detail.Arg1-as-detail \\Temporal.Synchronous} \\ \hline
  B in case A. & $A\to B$ & Contingency.Condition.Arg2-as-cond \\ \hline
  \makecell[l]{B. In fact, A. \\B, A in fact.} & $A\to B$ & \makecell[l]{Expansion.Instantiation.Arg2-as-instance \\Expansion.Level-of-detail.Arg1-as-detail \\Expansion.Level-of-detail.Arg2-as-detail} \\ \hline
  B in order A. & $A\to B$ & \makecell[l]{Contingency.Condition.Arg2-as-cond \\Contingency.Purpose.Arg2as-goal} \\ \hline
  B, in particular A. & $A\to B$ & \makecell[l]{Expansion.Instantiation.Arg2-as-instance \\Expansion.Level-of-detail.Arg2-as-detail} \\ \hline
  A, in short, B. & $A\to B$ & Expansion.Level-of-detail.Arg1-as-detail \\ \hline
  A, in sum, B. & $A\to B$ & Expansion.Level-of-detail.Arg1-as-detail \\ \hline
  B, in that A. & $A\to B$ & Expansion.Level-of-detail.Arg2-as-detail \\ \hline
  A, in the end B. & $A\to B$ & \makecell[l]{Contingency.Cause.Result \\Expansion.Level-of-detail.Arg1-as-detail \\Expansion.Level-of-detail.Arg2-as-detail \\Temporal.Asynchronous.Pre cedence} \\ \hline
  \makecell[l]{B, indeed A. \\B, A indeed.} & $A\to B$ & \makecell[l]{Contingency.Cause+Belief.Reason+Belief \\Contingency.Cause.Reason \\Contingency.Cause.Result \\Expansion.Conjunction \\Expansion.Instantiation.Arg2-as-instance \\Expansion.Level-of-detail.Arg1-as-detail \\Expansion.Level-of-detail.Arg2-as-detail} \\ \hline
  \makecell[l]{B insofar as A. \\Insofar as A, B.} & $A\to B$ & \makecell[l]{Contingency.Cause.Reason \\Expansion.Level-of-detail.Arg2-as-detail} \\ \hline
  A, B later. & $A\to B$ & Temporal.Asynchronous.Precedence \\ \hline
  A, B later on. & $A\to B$ & Temporal.Asynchronous.Precedence \\ \hline
  B, more accurately, A. & $A\to B$ & Expansion.Substitution.Arg2-as-subst \\ \hline
  A, next B. & $A\to B$ & \makecell[l]{Temporal.Asynchronous.Precedence \\Expansion.Conjunction \\Temporal.Asynchronous.Precedence} \\ \hline
  B, not only because of A. & $A\to B$ & Contingency.Cause.Reason \\ \hline
  Now that A, B. & $A\to B$ & \makecell[l]{Contingency.Cause.Reason \\Temporal.Asynchronous.Precedence  \\Contingency.Cause.Reason \\Temporal.Asynchronous.Succession \\Contingency.Cause.Reason \\Temporal.Synchronous \\Contingency.Cause.Reason \\Temporal.Synchronous} \\ \hline
  B on A. & $A\to B$ & Contingency.Cause.Reason \\ \hline
  \makecell[l]{Once A, B. \\B, once A.} & $A\to B$ & \makecell[l]{Contingency.Condition.Arg2-as-cond \\Temporal.Asynchronous.Succession \\Contingency.Cause.Reason \\Temporal.Asynchronous.Succession \\Contingency.Condition.Arg2as-cond \\Temporal.Asynchronous.Succession} \\ \hline
  B only if A. & $A\to B$ & Contingency.Condition.Arg2-as-cond \\ \hline
  B previously A. & $A\to B$ & \makecell[l]{Temporal.Asynchronous.Succession \\Comparison.Contrast \\Temporal.Asynchronous.Succession} \\ \hline
  \makecell[l]{B, since A. \\Since A, B.} & $A\to B$ & \makecell[l]{Contingency.Cause.Reason \\Temporal.Asynchronous.Precedence \\Temporal.Asynchronous.Succession \\Contingency.Cause.Reason \\Temporal.Asynchronous.Succession} \\ \hline
  B since before A. & $A\to B$ & Temporal.Asynchronous.Succession \\ \hline
  A, so B. & $A\to B$ & \makecell[l]{Contingency.Cause+Belief.Result+Belief \\Contingency.Cause.Result \\Contingency.Purpose.Arg2as-goal} \\ \hline
  So as A, B. & $A\to B$ & Contingency.Purpose.Arg2as-goal \\ \hline
  \makecell[l]{B so long as A. \\So long as A, B.} & $A\to B$ & Contingency.Condition.Arg2-as-cond \\ \hline
  A so that B. & $A\to B$ & \makecell[l]{Contingency.Cause.Result \\Contingency.Purpose.Arg2as-goal} \\ \hline
  B, specifically, A. & $A\to B$ & Expansion.Level-of-detail.Arg2-as-detail \\ \hline
  A subsequently B. & $A\to B$ & Temporal.Asynchronous.Precedence \\ \hline
  B such as A. & $A\to B$ & Expansion.Instantiation.Arg2-as-instance \\ \hline
  B, that is A. & $A\to B$ & \makecell[l]{Expansion.Equivalence \\Expansion.Level-of-detail.Arg2-as-detail} \\ \hline
  A then B. & $A\to B$ & \makecell[l]{Contingency.Cause.Result \\Expansion.Conjunction Contingency.Cause.Result \\Contingency.Condition.Arg1-as-cond \\Temporal.Asynchronous.Precedence \\Contingency.Cause.Result \\Temporal.Asynchronous.Precedence} \\ \hline
  A thereafter B. & $A\to B$ & Temporal.Asynchronous.Precedence \\ \hline
  A thereby B. & $A\to B$ & \makecell[l]{Contingency.Cause.Result \\Expansion.Manner.Arg1-asmanner} \\ \hline
  A therefore B. & $A\to B$ & Contingency.Cause.Result \\ \hline
  A thus B. & $A\to B$ & \makecell[l]{Contingency.Cause+Belief.Result+Belief \\Contingency.Cause.Result} \\ \hline
  B till A. & $A\to B$ & \makecell[l]{Contingency.Negative-condition.Arg2-as-negCond \\Temporal.Asynchronous.Precedence} \\ \hline
  A ultimately B. & $A\to B$ & \makecell[l]{Contingency.Cause.Reason \\Expansion.Conjunction \\Temporal.Asynchronous.Precedence \\Contingency.Cause.Result \\Temporal.Asynchronous.Precedence} \\ \hline
  \makecell[l]{B untill A. \\Untill A, B.} & $A\to B$ & \makecell[l]{Contingency.Condition.Arg2-as-cond \\Temporal.Asynchronous.Precedence \\Temporal.Asynchronous.Succession \\Contingency.Condition.Arg2-as-cond \\Temporal.Asynchronous.Su ccession} \\ \hline
  \makecell[l]{B upon A. \\Upon A, B.} & $A\to B$ & \makecell[l]{Temporal.Asynchronous.Succession \\Contingency.Cause.Reason \\Temporal.Synchronous \\Contingency.Cause.Reason \\Temporal.Synchronous} \\ \hline
  \makecell[l]{B, when A. \\When A, B.} & $A\to B$ & \makecell[l]{Contingency.Cause.Reason \\Contingency.Condition+SpeechAct \\Contingency.Condition.Arg2-as-cond \\Expansion.Level-of-detail.Arg2-as-detail \\Contingency.Condition.Arg2-as-cond \\Expansion.Manner.Arg2-asmanner \\Temporal.Asynchronous.Precedence \\Contingency.Condition.Arg2-as-cond \\Temporal.Asynchronous.Precedence \\Temporal.Asynchronous.Succession \\Contingency.Cause+Belief.Reason+Belief \\Temporal.Asynchronous.Succession \\Contingency.Cause.Reason \\Temporal.Asynchronous.Succession \\Contingency.Cause.Result \\Temporal.Asynchronous.Succession \\Contingency.Condition+SpeechAct \\Temporal.Asynchronous.Succession \\Contingency.Condition.Arg2-as-cond \\Temporal.Asynchronous.Su ccession} \\ \hline
  B when and if A. & $A\to B$ & \makecell[l]{Temporal.Asynchronous.Succession \\Contingency.Condition.Arg2-as-cond} \\ \hline
  \makecell[l]{B whenever A. \\Whenever A, B.} & $A\to B$ & Contingency.Condition.Arg2-as-cond \\ \hline
  \makecell[l]{B, where A. \\Where A, B.} & $A\to B$ & Contingency.Condition.Arg2-as-cond \\ \hline
  \makecell[l]{B, with A. \\With A, B} & $A\to B$ & \makecell[l]{Contingency.Cause+Belief.Reason+Belief \\Contingency.Cause.Reason \\Expansion.Level-of-detail.Arg2-as-detail \\Contingency.Cause.Reason \\Contingency.Condition.Arg2-as-cond \\Expansion.Instantiation.Arg2-as-instance \\Expansion.Level-of-detail.Arg2-as-detail \\Expansion.Manner.Arg2-asmanner} \\ \hline
  \makecell[l]{B, without A. \\Without A, B.} & $A\to B$ & \makecell[l]{Contingency.Cause.Reason \\Contingency.Cause.Result \\Expansion.Level-of-detail.Arg2-as-detail \\Expansion.Manner.Arg2-asmanner} \\ 

  \bottomrule
  \caption{
    \label{tab:pdtb_implication}
    Mapping from connectives to logical implication ($\to$), according to PDTB senses. 
  }
\end{longtable}  

\newpage
\begin{longtable}{  
    lcl
  }
  \toprule
  \textbf{Connectives} & \textbf{Binary Operator} & \textbf{Senses} \\
  \midrule
  A, and B. & $A\land B$ & \makecell[l]{Comparison.Concession+SpeechAct.Arg2-as-denier+SpeechAct \\Comparison.Contrast \\Contingency.Cause+SpeechAct.Result+SpeechAct \\Contingency.Cause.Reason \\Contingency.Cause.Result \\Expansion.Conjunction \\Contingency.Cause.Result \\Contingency.Condition.Arg1-as-cond \\Contingency.Purpose.Arg2as-goal \\Expansion.Conjunction \\Expansion.Level-of-detail.Arg2-as-detail \\Expansion.Manner.Arg2-asmanner} \\ \hline
  additionally & $A\land B$ & Expansion.Conjunction \\ \hline
  \makecell[l]{Albeit A, B. \\B, albeit A.} & $A\land B$ & Comparison.Concession.Arg2-as-denier \\ \hline
  along with & $A\land B$ & Expansion.Conjunction \\ \hline
  also & $A\land B$ & \makecell[l]{Expansion.Conjunction \\Temporal.Synchronous} \\ \hline
  although & $A\land B$ & \makecell[l]{Comparison.Concession.Arg1-as-denier \\Comparison.Concession.Arg2-as-denier \\Comparison.Contrast \\Expansion.Exception.Arg2-as-excpt \\Temporal.Synchronous \\Comparison.Contrast} \\ \hline
  as long as & $A\land B$ & \makecell[l]{Temporal.Synchronous \\Contingency.Condition.Arg2-as-cond \\Temporal.Synchronous} \\ \hline
  as much as & $A\land B$ & \makecell[l]{Comparison.Concession.Arg1-as-denier \\Expansion.Conjunction \\Expansion.Substitution.Arg2-as-subst} \\ \hline
  as soon as & $A\land B$ & \makecell[l]{Temporal.Asynchronous.Succession \\Temporal.Synchronous} \\ \hline
  as though & $A\land B$ & \makecell[l]{Comparison.Similarity \\Expansion.Manner.Arg2-asmanner \\Comparison.Similarity \\Expansion.Level-of-detail.Arg2-as-detail} \\ \hline
  as well & $A\land B$ & \makecell[l]{Comparison.Similarity \\Expansion.Conjunction} \\ \hline
  as well as & $A\land B$ & \makecell[l]{Expansion.Conjunction} \\ \hline
  as & $A\land B$ & \makecell[l]{Comparison.Concession.Arg1-as-denier \\Comparison.Contrast \\Comparison.Similarity \\Temporal.Synchronous \\Comparison.Contrast \\Temporal.Synchronous \\Comparison.Similarity \\Temporal.Synchronous \\Contingency.Cause+Belief.Reason+Belief \\Temporal.Synchronous \\Contingency.Cause.Reason \\Temporal.Synchronous} \\ \hline
  at the same time & $A\land B$ & \makecell[l]{Temporal.Synchronous} \\ \hline
  before and after & $A\land B$ & \makecell[l]{Temporal.Asynchronous.Precedence \\Temporal.Asynchronous.Succession} \\ \hline
  besides & $A\land B$ & Expansion.Conjunction \\ \hline
  \makecell[l]{A, beyond B. \\Beyond B, A.} & $A\land B$ & Expansion.Conjunction \\ \hline
  both A and B. & $A\land B$ & Expansion.Conjunction \\ \hline
  but & $A\land B$ & \makecell[l]{Comparison.Concession+SpeechAct.Arg2-as-denier+SpeechAct \\Comparison.Concession.Arg2-as-denier \\Comparison.Contrast \\Contingency.Cause+SpeechAct.Reason+SpeechAct \\Contingency.Cause.Reason \\Comparison.Concession.Arg2-as-denier \\Expansion.Conjunction \\Expansion.Exception.Arg2-as-excpt \\Temporal.Synchronous \\Comparison.Contrast} \\ \hline
  A but also B. & $A\land B$ & Expansion.Conjunction \\ \hline
  A but then B. & $A\land B$ & Comparison.Concession.Arg2-as-denier \\ \hline
  A but then again B. & $A\land B$ & Comparison.Concession.Arg2-as-denier \\ \hline
  by comparison & $A\land B$ & Comparison.Contrast \\ \hline
  by contrast & $A\land B$ & Comparison.Contrast \\ \hline
  conversely & $A\land B$ & Comparison.Contrast \\ \hline
  \makecell[l]{Despite A, B. \\B, despite A.} & $A\land B$ & Comparison.Concession.Arg2-as-denier \\ \hline
  \makecell[l]{A even as B. \\Even as B, A.} & $A\land B$ & \makecell[l]{Comparison.Concession.Arg1-as-denier \\Temporal.Synchronous \\Comparison.Concession.Arg1-as-denier} \\ \hline
  even before & $A\land B$ & \makecell[l]{Temporal.Asynchronous.Precedence \\Comparison.Concession.Arg1-as-denier} \\ \hline
  even before then & $A\land B$ & \makecell[l]{Temporal.Asynchronous.Succession \\Comparison.Concession.Arg2-as-denier} \\ \hline
  even if & $A\land B$ & Comparison.Concession.Arg1-as-denier \\ \hline
  even so & $A\land B$ & Comparison.Concession.Arg2-as-denier \\ \hline
  even then & $A\land B$ & \makecell[l]{Temporal.Asynchronous.Precedence \\Comparison.Concession.Arg2-as-denier} \\ \hline
  even though & $A\land B$ & \makecell[l]{Comparison.Concession.Arg1-as-denier \\Comparison.Concession.Arg2-as-denier} \\ \hline
  even when & $A\land B$ & \makecell[l]{Comparison.Concession.Arg1-as-denier \\Temporal.Asynchronous.Succession \\Comparison.Concession.Arg1-as-denier \\Temporal.Synchronous \\Comparison.Concession.Arg1-as-denier} \\ \hline
  even while & $A\land B$ & \makecell[l]{Temporal.Synchronous \\Comparison.Concession.Arg1-as-denier} \\ \hline
  even with & $A\land B$ & Comparison.Concession.Arg1-as-denier \\ \hline
  finally & $A\land B$ & Expansion.Conjunction \\ \hline
  further & $A\land B$ & Expansion.Conjunction \\ \hline
  furthermore & $A\land B$ & Expansion.Conjunction \\ \hline
  A however B. & $A\land B$ & \makecell[l]{Comparison.Concession.Arg1-as-denier \\Comparison.Concession.Arg2-as-denier \\Comparison.Contrast \\Temporal.Synchronous \\Comparison.Contrast} \\ \hline
  in addition & $A\land B$ & Expansion.Conjunction \\ \hline
  in any case & $A\land B$ & Comparison.Concession.Arg2-as-denier \\ \hline
  in contrast & $A\land B$ & Comparison.Contrast \\ \hline
  in fact & $A\land B$ & \makecell[l]{Comparison.Concession.Arg2-as-denier \\Comparison.Contrast \\Expansion.Conjunction} \\ \hline
  in the end & $A\land B$ & \makecell[l]{Comparison.Concession.Arg2-as-denier \\Comparison.Contrast \\Expansion.Conjunction} \\ \hline
  in the meantime & $A\land B$ & \makecell[l]{Temporal.Asynchronous.Succession \\Temporal.Synchronous \\Comparison.Contrast \\Temporal.Synchronous} \\ \hline
  in the meanwhile & $A\land B$ & Temporal.Synchronous \\ \hline
  indeed & $A\land B$ & \makecell[l]{Comparison.Concession.Arg2-as-denier \\Expansion.Conjunction \\Expansion.Equivalence} \\ \hline
  like & $A\land B$ & \makecell[l]{Comparison.Contrast \\Comparison.Similarity \\Expansion.Instantiation.Arg2-as-instance} \\ \hline
  likewise & $A\land B$ & Expansion.Conjunction \\ \hline
  meantime & $A\land B$ & Temporal.Synchronous \\ \hline
  meanwhile & $A\land B$ & \makecell[l]{Comparison.Concession.Arg2-as-denier \\Comparison.Contrast \\Expansion.Conjunction \\Temporal.Synchronous \\Comparison.Concession.Arg2-as-denier \\Temporal.Synchronous \\Comparison.Contrast \\Temporal.Synchronous \\Comparison.Similarity \\Temporal.Synchronous} \\ \hline
  moreover & $A\land B$ & Expansion.Conjunction \\ \hline
  much less & $A\land B$ & Expansion.Conjunction \\ \hline
  neither A nor B. & $A\land B$ & \makecell[l]{Comparison.Contrast \\Expansion.Conjunction} \\ \hline
  nevertheless & $A\land B$ & \makecell[l]{Comparison.Concession.Arg2-as-denier \\Comparison.Contrast} \\ \hline
  no matter & $A\land B$ & Comparison.Concession.Arg1-as-denier \\ \hline
  nonetheless & $A\land B$ & \makecell[l]{Comparison.Concession.Arg2-as-denier \\Comparison.Contrast} \\ \hline
  A; nor B. & $A\land B$ & \makecell[l]{Comparison.Concession.Arg2-as-denier \\Expansion.Conjunction} \\ \hline
  not just A, but B. & $A\land B$ & \makecell[l]{Comparison.Contrast \\Expansion.Conjunction} \\ \hline
  not just A, but also B. & $A\land B$ & \makecell[l]{Comparison.Contrast \\Expansion.Conjunction} \\ \hline
  not only & $A\land B$ & \makecell[l]{Comparison.Contrast \\Expansion.Conjunction} \\ \hline
  not only A, also B. & $A\land B$ & \makecell[l]{Comparison.Contrast \\Expansion.Conjunction} \\ \hline
  not only A but B. & $A\land B$ & \makecell[l]{Comparison.Concession.Arg2-as-denier \\Expansion.Conjunction} \\ \hline
  not only A but also B. & $A\land B$ & \makecell[l]{Comparison.Contrast \\Expansion.Conjunction} \\ \hline
  on the contrary & $A\land B$ & Comparison.Contrast \\ \hline
  \makecell[l]{on the one hand A \\on the other B.} & $A\land B$ & Comparison.Contrast \\ \hline
  \makecell[l]{on the one hand A \\on the other hand B.} & $A\land B$ & \makecell[l]{Comparison.Concession.Arg2-as-denier \\Comparison.Contrast} \\ \hline
  on the other hand & $A\land B$ & \makecell[l]{Comparison.Concession.Arg2-as-denier \\Comparison.Contrast} \\ \hline
  only & $A\land B$ & \makecell[l]{Comparison.Concession.Arg2-as-denier \\Comparison.Contrast \\Expansion.Exception.Arg2-as-excpt \\Expansion.Level-of-detail.Arg2-as-detail} \\ \hline
  A or B. & $A\land B$ & \makecell[l]{Comparison.Concession+SpeechAct.Arg2-as-denier+SpeechAct \\Comparison.Concession.Arg2-as-denier \\Contingency.Condition+SpeechAct \\Contingency.Negative-condition.Arg1-as-negCond \\Expansion.Conjunction \\Expansion.Equivalence} \\ \hline
  plus & $A\land B$ & Expansion.Conjunction \\ \hline
  regardless & $A\land B$ & Comparison.Concession.Arg2-as-denier \\ \hline
  regardless of & $A\land B$ & Comparison.Concession.Arg1-as-denier \\ \hline
  separately & $A\land B$ & \makecell[l]{Expansion.Conjunction \\Temporal.Synchronous \\Expansion.Conjunction} \\ \hline
  similarly & $A\land B$ & Comparison.Similarity \\ \hline
  simultaneously & $A\land B$ & Temporal.Synchronous \\ \hline
  still & $A\land B$ & \makecell[l]{Comparison.Concession.Arg2-as-denier \\Comparison.Contrast \\Temporal.Asynchronous.Precedence \\Temporal.Synchronous} \\ \hline
  A then B. & $A\land B$ & \makecell[l]{Expansion.Conjunction \\Temporal.Synchronous} \\ \hline
  though & $A\land B$ & \makecell[l]{Comparison.Concession.Arg1-as-denier \\Comparison.Concession.Arg2-as-denier \\Comparison.Contrast} \\ \hline
  A whatever B. & $A\land B$ & Comparison.Concession.Arg1-as-denier \\ \hline
  when & $A\land B$ & \makecell[l]{Comparison.Concession.Arg1-as-denier \\Comparison.Concession.Arg2-as-denier \\Comparison.Contrast \\Temporal.Synchronous \\Comparison.Contrast \\Temporal.Synchronous \\Contingency.Cause+Belief.Reason+Belief \\Temporal.Synchronous \\Contingency.Cause.Reason \\Temporal.Synchronous \\Contingency.Cause.Result \\Temporal.Synchronous \\Contingency.Condition+SpeechAct \\Temporal.Synchronous \\Contingency.Condition.Arg2-as-cond \\Temporal.Synchronous \\Expansion.Level-of-detail.Arg2-as-detail \\Temporal.Synchronous} \\ \hline
  whereas & $A\land B$ & Comparison.Contrast \\ \hline
  whether & $A\land B$ & Comparison.Concession.Arg1-as-denier \\ \hline
  while & $A\land B$ & \makecell[l]{Comparison.Concession.Arg1-as-denier \\Comparison.Concession.Arg2-as-denier \\Comparison.Contrast \\Comparison.Similarity \\Expansion.Conjunction \\Temporal.Synchronous \\Comparison.Concession.Arg1-as-denier \\Temporal.Synchronous \\Comparison.Concession.Arg2-as-denier \\Temporal.Synchronous \\Comparison.Contrast \\Temporal.Synchronous \\Expansion.Conjunction \\Temporal.Synchronous} \\ \hline
  with & $A\land B$ & \makecell[l]{Comparison.Concession.Arg1-as-denier \\Comparison.Contrast \\Expansion.Conjunction \\Temporal.Synchronous} \\ \hline
  yet & $A\land B$ & \makecell[l]{Comparison.Concession.Arg2-as-denier \\Comparison.Contrast \\Expansion.Conjunction} \\ 
  \bottomrule
  \caption{
    \label{tab:pdtb_and}
    Mapping from connectives to logical conjunction ($\land$), according to PDTB senses. 
  }
\end{longtable}

\newpage
\begin{longtable}{  
    lcl
  }
  \toprule
  \textbf{Connectives} & \textbf{Binary Operator} & \textbf{Senses} \\
  \midrule
  alternatively & $A\lor B$ & \makecell[l]{Expansion.Disjunction \\Expansion.Substitution.Arg2-as-subst} \\ \hline
  as an alternative & $A\lor B$ & Expansion.Disjunction \\ \hline
  either A or B. & $A\lor B$ & \makecell[l]{Contingency.Negative-condition.Arg1-as-negCond \\Expansion.Disjunction} \\ \hline
  A except B. & $A\lor B$ & Expansion.Exception.Arg2-as-excpt \\ \hline
  in other words & $A\lor B$ & Expansion.Equivalence \\ \hline
  instead & $A\lor B$ & Expansion.Substitution.Arg2-as-subst \\ \hline
  instead of & $A\lor B$ & Expansion.Substitution.Arg1-as-subst \\ \hline
  \makecell[l]{A, lest B. \\Lest B, A.} & $A\lor B$ & Contingency.Negative-condition.Arg1-as-negCond \\ \hline
  not so much as & $A\lor B$ & Expansion.Substitution.Arg2-as-subst \\ \hline
  A, or B. & $A\lor B$ & Expansion.Disjunction \\ \hline
  or otherwise & $A\lor B$ & Expansion.Disjunction \\ \hline
  otherwise & $A\lor B$ & \makecell[l]{Contingency.Negative-condition.Arg1-as-negCond \\Expansion.Exception.Arg1-as-excpt} \\ \hline
  rather & $A\lor B$ & Expansion.Substitution.Arg2-as-subst \\ \hline
  rather than & $A\lor B$ & Expansion.Substitution.Arg1-as-subst \\ \hline
  so much as & $A\lor B$ & Expansion.Substitution.Arg2-as-subst \\ \hline
  \makecell[l]{A unless B. \\Unless B, A.} & $A\lor B$ & Contingency.Negative-condition.Arg2-as-negCond \\ \hline
  A, without B & $A\lor B$ & Contingency.Negative-condition.Arg2-as-negCond \\
  \bottomrule
  \caption{
    \label{tab:pdtb_or}
    Mapping from connectives to logical disjunction ($\lor$), according to PDTB senses. 
  }
\end{longtable}



\end{document}